\documentclass[%
 aip,
 amsmath,amssymb,
 reprint,%
 floatfix,%
]{revtex4-1}

\usepackage{graphicx}
\usepackage{adjustbox}
\usepackage{dcolumn}
\usepackage{bm}
\usepackage{booktabs}
\usepackage{multirow}
\usepackage[utf8]{inputenc}
\usepackage[T1]{fontenc}
\usepackage{mathptmx}
\usepackage{etoolbox}
\usepackage{subcaption}
\usepackage{hyperref}
\usepackage[table]{xcolor}
\usepackage{placeins}
\usepackage{tikz}
\usetikzlibrary{positioning,arrows.meta,fit,backgrounds,calc}
\usepackage{algpseudocode}
\usepackage{makecell}
\usepackage{xspace}
\usepackage{listings}
\usepackage[capitalize,noabbrev]{cleveref}

\DeclareCaptionType{algorithm}[Algorithm][List of Algorithms]
\crefname{algorithm}{Algorithm}{Algorithms}
\Crefname{algorithm}{Algorithm}{Algorithms}

\lstdefinestyle{mpistyle}{%
  basicstyle=\ttfamily\footnotesize,
  keywordstyle=\bfseries,
  commentstyle=\itshape\color{gray!70!black},
  showstringspaces=false,
  columns=fullflexible,
  keepspaces=true,
  breaklines=true,
  frame=single,
  framesep=4pt,
  rulecolor=\color{gray!55},
  aboveskip=4pt,
  belowskip=2pt,
  xleftmargin=2pt,
  language=Python,
}

\newcommand{\dblfigwide}{0.92\textwidth}
\newcommand{\dblfighalf}{0.48\textwidth}

\newcommand{\mpinode}{MPINeuralODE\xspace}

\makeatletter
\def\@email#1#2{%
 \endgroup
 \patchcmd{\titleblock@produce}
  {\frontmatter@RRAPformat}
  {\frontmatter@RRAPformat{\produce@RRAP{*#1\href{mailto:#2}{#2}}}\frontmatter@RRAPformat}
  {}{}
}%
\makeatother

\begin{document}

\preprint{AIP/123-QED}

\title[Physics-Informed Neural ODEs for Dynamical Systems]{\mpinode: Multiple-Initial-Condition Physics-Informed Neural ODEs for Globally Consistent Dynamical System Learning}

\author{Lake Yang}
 \email{l.yang1@imperial.ac.uk}
 \homepage{https://www.linkedin.com/in/lake-yang-8924b51b7}
\author{Antonio Malpica-Morales}
\author{Frank Ioannis Papadakis Wood}
\author{Serafim Kalliadasis}
\affiliation{%
Department of Chemical Engineering, Imperial College London, London SW7 2AZ, United Kingdom
}

\date{\today}

\begin{abstract}
A Neural ordinary differential equation (Neural ODE) learns a
system's dynamics from data, but the learned vector field is
typically faithful only near the training trajectories: from an
unseen initial condition it can spiral into unphysical states, and
over long horizons it dissipates invariants the true dynamics
conserve. We propose \mpinode, which augments a Neural ODE with two
complementary ingredients that recover the dynamics \emph{globally}
--- across phase space, not just along the observed paths --- so that
the model extrapolates to unseen initial conditions and stays
faithful over long forecasts. The first ingredient is a soft
physics-informed residual; the second is a Multiple-Initial-Condition
(MIC) multiple-shooting curriculum. The two are structurally
complementary rather than additive: the physics term anchors the
magnitude of the learned vector field on precisely the region of
phase space that the MIC sampler enlarges, while the MIC sampler in
turn supplies the broad support on which the physics residual can be
meaningfully estimated. We make this closure quantitative with an
elementary Gr\"onwall--Lipschitz bound that factors the prediction
error into a horizon-growth term and a residual-magnitude term,
controlled by the two ingredients separately, and we argue that a
single scalar trajectory error cannot reveal whether a model has
recovered the true vector field. We therefore evaluate every model
along three complementary axes --- out-of-sample error, long-horizon
stability, and Hamiltonian drift. On the Lotka--Volterra benchmark,
\mpinode\ achieves the lowest out-of-sample and long-horizon
mean-squared error (MSE) among data-driven methods (a 24\% reduction
relative to the baseline Neural ODE) and the lowest Hamiltonian
drift, making it the strongest method across all three axes, using a
lightweight ($\sim\!67$k-parameter) vector-field network and no
knowledge of the governing equations. The construction is
system-agnostic and drops into an existing Neural ODE as a
``surgery'' upgrade; we release it as an installable Python package.
\end{abstract}

\maketitle

\section{Introduction}
\label{sec:introduction}

Dynamical systems describe how state variables evolve under internal
interactions and external forcing, from predator--prey ecology%
~\cite{Murray2002Mathematical} to chemical kinetics and geophysical
fluid flow. A persistent tension in data-driven modeling is the
tradeoff between mechanistic interpretability and empirical
flexibility. Classical mechanistic models~\cite{Strogatz2015Nonlinear}
provide identifiable rate constants but depend on assumptions
--- well-mixed populations, linear functional responses, absence of
unmodeled dynamics --- that are routinely violated in practice.
Purely data-driven approaches~\cite{Voss2004Nonlinear,Brunton2016SINDy,Pathak2018Reservoir}
offer representational capacity but sacrifice interpretability and
often generalize poorly outside the training distribution.

\begin{figure*}[tbp]
\centering
\begin{subfigure}[t]{\dblfighalf}
  \includegraphics[width=\linewidth]{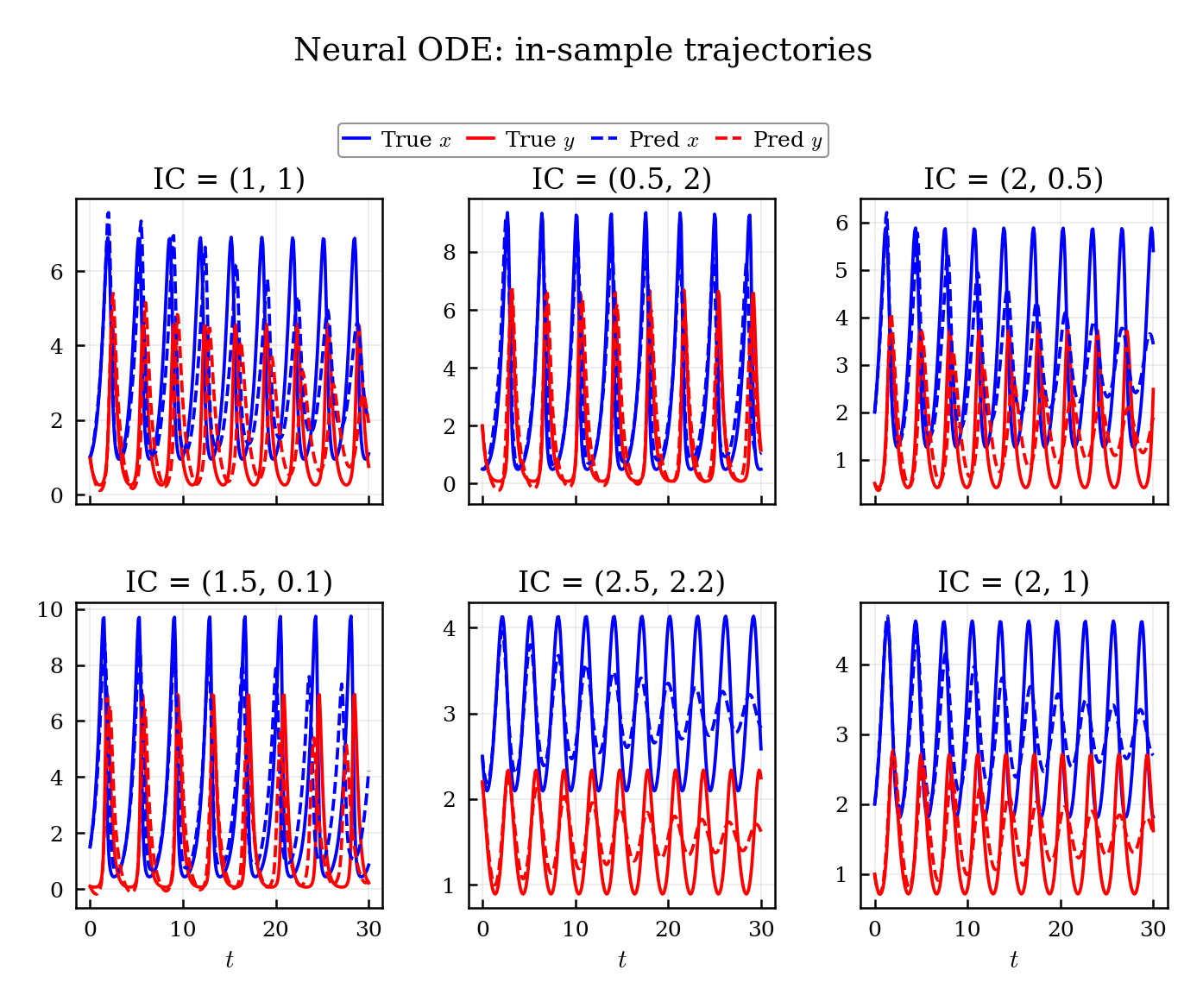}
  \caption{In-sample time series: the baseline neural network (NN) fits training trajectories accurately across all six panels.}
\end{subfigure}\hfill
\begin{subfigure}[t]{0.40\textwidth}
  \includegraphics[width=\linewidth]{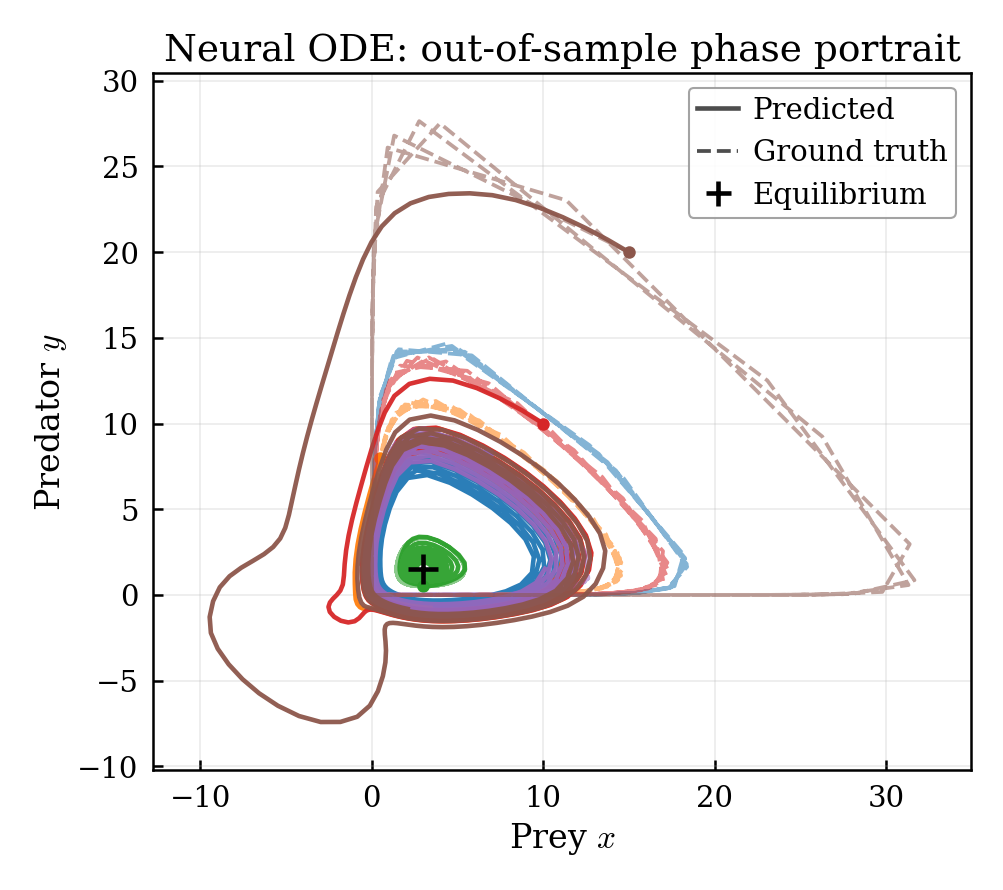}
  \caption{Out-of-sample phase portrait: predicted orbits (solid lines) spiral and cross axes rather than tracing the ground-truth closed orbits (dashed lines).}
\end{subfigure}
\caption{\textbf{Baseline Neural ODE failure modes on Lotka--Volterra.}
Accurate in-sample fit \textbf{(a)} coexists with structurally
broken out-of-sample behavior \textbf{(b)}: the learned vector field
is valid only in the training corridor and produces qualitatively
wrong dynamics elsewhere. The closed-orbit topology of the true
system is replaced by artificial spirals. In all panels predicted
trajectories are drawn solid and ground-truth trajectories dashed,
so that the two are distinguishable independently of color.}
\label{fig:neural_ode_failure}
\end{figure*}

Neural ordinary differential equations (Neural ODEs)~\cite{Chen2018NODE}
parameterize the instantaneous vector field with a neural network
and obtain trajectories by numerical integration. The formulation
handles irregularly-sampled data~\cite{Rubanova2019Latent}, produces
smooth solver-controlled trajectories, and supports training through
the adjoint method~\cite{Bradley2013Adjoint}. Yet our experiments
confirm earlier observations~\cite{Krishnapriyan2021Characterizing,Wang2021Understanding}
that pure Neural ODEs can learn vector fields valid only near
training trajectories. \Cref{fig:neural_ode_failure} illustrates
this: on the Lotka--Volterra system, the baseline Neural ODE fits
training trajectories well, but on held-out initial conditions it
produces artificial spirals where the true dynamics admit
conservative closed orbits.

This paper investigates two complementary remedies and their
combination. Soft \emph{physics-informed regularization}%
~\cite{Raissi2017PhysicsI,Raissi2019PINN} steers the learned vector
field toward physically plausible behavior at sampled collocation
points; this construction shares its variational structure with the
classical optimal-control formulations of Pontryagin and
Bellman~\cite{Pontryagin1962Mathematical,Sidhoum2025PDDP}.
\emph{Multiple-Initial-Condition multiple-shooting}%
~\cite{Bock1984Multiple,Turan2022Multiple} broadens phase-space
coverage and enforces continuity of the learned flow across
segments. A central methodological observation is that a single
scalar metric --- such as validation MSE --- does not reveal whether
a learned model has recovered the true dynamics or merely fit
observed trajectories. We evaluate every method on three
complementary axes:

\begin{enumerate}\setlength\itemsep{1pt}
\item \textbf{Out-of-sample (OOS) prediction error} --- accuracy on
      held-out initial conditions in the training distribution.
\item \textbf{Long-horizon stability} --- accumulated error over
      multiple oscillation periods, exposing compounding error.
\item \textbf{Drift in a conserved quantity} (here the Hamiltonian
      $H$) --- a trajectory-independent measure of whether the
      learned dynamics respect the geometric structure of phase
      space, applicable to any system that admits such an invariant.
\end{enumerate}

A model that performs well on all three has captured something close
to the true vector field; a model that excels only on OOS error may
still drift over long horizons or dissipate invariants.

\paragraph{Contributions.}
(1)~We propose \mpinode, a Neural ODE framework whose physics
residual and MIC continuity penalty are designed as a closure
relation on the visited support of phase space rather than a
weighted sum of independent objectives: each ingredient supplies
exactly the inductive bias the other is structurally missing. We make
this closure quantitative with a Gr\"onwall--Lipschitz bound
(\cref{eq:gronwall}) that factors the prediction error into a
horizon-growth term and a residual-magnitude term, controlled by the
two ingredients separately.
(2)~We argue that single-scalar trajectory error is insufficient for
evaluating learned dynamical systems and propose a three-axis report
that exposes failure modes a scalar MSE hides.
(3)~On the Lotka--Volterra benchmark, \mpinode\ achieves the lowest
OOS and long-horizon error among purely data-driven methods, and
the lowest Hamiltonian drift --- the strongest method across all
three axes. No axis is worse than any single ingredient alone. In the
typical practitioner regime where exact equations are unavailable,
this makes it the most practical surrogate among the methods we
evaluated, using only a lightweight ($\sim\!67$k-parameter)
vector-field network and no knowledge of the governing law.
(4)~The construction is system-agnostic and drops into an existing
Neural ODE as a ``surgery'' upgrade rather than a from-scratch
rewrite. We release \mpinode\ as an installable Python package that
provides a ready-to-use \texttt{LotkaVolterra} implementation
together with a documented base-class interface for defining new
dynamical systems.

\FloatBarrier

\section{Related Work}
\label{sec:related-work}

\paragraph{Neural ODEs and variants.}
Neural ODEs~\cite{Chen2018NODE} parameterize a continuous-time
vector field with a neural network. Latent
ODEs~\cite{Rubanova2019Latent}, Augmented Neural
ODEs~\cite{Dupont2019Augmented}, and Neural Controlled Differential
Equations~\cite{Kidger2020Neural} address architectural limitations
(partial observability, expressivity, exogenous control) but not
global consistency across initial conditions, which is our target.

\paragraph{Physics-informed losses.}
Physics-Informed Neural Networks (PINNs)~\cite{Raissi2017PhysicsI,Raissi2019PINN}
enforce soft constraints through collocation sampling but can
require extensive sampling and struggle with long-horizon
prediction~\cite{Krishnapriyan2021Characterizing,Wang2021Understanding}.
We integrate the PINN idea into the Neural ODE framework, reducing
collocation requirements by sampling along the integrated trajectory
rather than over a phase-space grid. The same variational logic that
makes a residual a useful constraint at sampled collocation points
also underlies the classical trajectory-optimization framework of
Pontryagin's minimum principle and Bellman's dynamic
programming~\cite{Pontryagin1962Mathematical,Sidhoum2025PDDP}; both
are statements about how local pointwise conditions propagate to
trajectory-level optimality. Our use of a Monte-Carlo collocation
estimator is the soft-constraint analog of that pointwise principle.

\paragraph{Multiple shooting.}
The direct multiple-shooting method originated in numerical optimal
control~\cite{Bock1984Multiple} to overcome the phase-mismatch
pathology of single-shooting on long-horizon oscillatory systems.
Turan and J\"aschke~\cite{Turan2022Multiple} first applied this
idea to neural differential equations, showing it can rescue the
``flattened-out'' trajectories single-shooting produces on
oscillatory time series. Our MIC mechanism is a multiple-shooting
curriculum in that lineage, augmented with a freshly-sampled
initial-condition distribution each epoch and combined with a
PINN-style residual that these works do not address.

\paragraph{Structure-preserving networks.}
Hamiltonian Neural Networks~\cite{Greydanus2019Hamiltonian}, related
Hamiltonian-system identification~\cite{Bertalan2019Inference}, and
symplectic networks~\cite{Jin2020SympNets} build conservation laws
into the architecture itself; this approach is orthogonal to and
potentially combinable with ours.

\paragraph{Mechanistic-neural hybrids.}
Universal Differential Equations (UDEs)~\cite{Rackauckas2020UDE}
embed a mechanistic model with a neural residual, achieving strong
performance when mechanistic knowledge is complete; we use this
construction as a \emph{fully-mechanistic upper-bound benchmark}
(see \cref{sec:ude}). Data-driven discovery
methods~\cite{Brunton2016SINDy,Champion2019DataDriven} excel when
the true model lies in a known function library but struggle when
the functional form is unknown --- the setting \mpinode\ targets.

\section{Computational Framework}
\label{sec:method}

\subsection{Lotka--Volterra Benchmark}
\label{sec:lv_setup}

The Lotka--Volterra (LV) system~\cite{Lotka1925Elements,Volterra1926Fluctuations}
\begin{equation}
\frac{dx}{dt} = \alpha x - \beta x y, \quad
\frac{dy}{dt} = \delta x y - \gamma y,
\label{eq:lotka_volterra}
\end{equation}
where $x(t)$ and $y(t)$ are prey and predator population densities
and $(\alpha,\beta,\gamma,\delta)$ are positive rate constants
(prey growth, predation, predator death, and predator reproduction
respectively), provides a mathematically tractable benchmark with
three useful properties. First, it exhibits neutrally-stable closed
orbits with no inherent error dissipation: the coexistence equilibrium
at $(\gamma/\delta,\,\alpha/\beta)$ has purely imaginary eigenvalues,
indicating a center~\cite{Strogatz2015Nonlinear}. Any artificial
damping or drift therefore accumulates rather than decays. Second,
families of orbits at different amplitudes share the same phase
plane, requiring the learner to generalize across orbits. Third,
the system admits the conserved quantity~\cite{Murray2002Mathematical}
\begin{equation}
H(x, y) = \delta\,x - \gamma \ln x + \beta\,y - \alpha \ln y,
\label{eq:hamiltonian}
\end{equation}
which we refer to as the Hamiltonian; it is constant along true
trajectories. Drift in $H(t)$ along a learned trajectory is
therefore a trajectory-independent diagnostic of how well the
model respects the geometric structure of phase space --- any
deviation from a constant is unambiguously a model error and does
not depend on which initial condition we picked. We adopt
$(\alpha, \beta, \gamma, \delta) = (1.5, 1.0, 3.0, 1.0)$, giving
oscillation period $T \approx 5.13$ and interior equilibrium
$(3.0, 1.5)$.

\paragraph{Data generation.}
Ground-truth trajectories are integrated with the adaptive
Dormand--Prince method~\cite{Dormand1980Runge} (the order $5(4)$
embedded Runge--Kutta pair, denoted \texttt{dopri5} in the
algorithms and tables below) at relative tolerance
$\textrm{rtol} = 10^{-8}$ and absolute tolerance
$\textrm{atol} = 10^{-8}$ (the integrator's two error-control
thresholds for accepting a step) over $t \in [0, 30]$ with sampling
step $\Delta t = 0.1$, yielding $301$ time points per trajectory
($\sim\!5.85$ oscillation cycles). Initial conditions (ICs) are
drawn from a mixed distribution: half uniform on $[0.1, 10.0]^2$
(typical regime) and half log-uniform on $[10^{-3}, 10.0]^2$ (edge
regime near the extinction axes). This mixed scheme avoids
pathological behavior near coordinate axes.

\subsection{Favoring a Neural ODE Configuration: An Experimental Campaign}
\label{sec:phase1}

Three implementation choices --- network architecture, temporal
sampling, and positivity wrapper --- are decided up front by an
experimental campaign of sequential ablations. Each ablation is
scored against a composite metric (the log-space mean of in-sample,
OOS, and long-horizon MSE) using only the baseline Neural ODE
variant, which decouples the choice from any physics-loss or MIC
effect. We use the term \emph{favored configuration} to denote the
configuration that achieves the lowest composite metric within
each ablation sweep, optimized by the Adam method~\cite{Kingma2014Adam};
this label is descriptive rather than absolute, and reflects the
specific composite metric, the validation set, and the seed budget
used here. The favored configurations of the campaign are summarized
in \cref{tab:phase1_winners}; per-variant diagnostics are reported
in \cref{app:architecture,app:temporal,app:positivity}.

\begin{table}[tbp]
\centering
\small
\caption{\textbf{Favored configurations from the Phase-1
experimental campaign.} Each row reports the variant that achieved
the lowest composite metric in its sweep. The positivity row reports
a tie between \emph{clamp} and \emph{none} on the composite metric;
we adopt clamp as a deterministic safety net.}
\label{tab:phase1_winners}
\setlength{\tabcolsep}{2pt}
\renewcommand{\arraystretch}{1.1}
\rowcolors{2}{gray!8}{white}
\begin{ruledtabular}
% NOTE: REVTeX's `ruledtabular' parses the column spec with its own
% \@testpach, which chokes on `p{}' (paragraph) columns and on the
% array-package `>{}' prefix -- either one triggers the fatal
% "Extra \or" / runaway-argument crash seen on arXiv. The cell
% contents here are short, so plain l/l columns suffice and are
% fully compatible with ruledtabular.
\begin{tabular}{lll}
Choice & Favored configuration & Variants compared \\
\colrule
Architecture      & $128 \times 4$ tanh  & widths $\{32,\dots,256\}$, depths $\{2,\dots,6\}$ \\
\addlinespace[1pt]
Temporal sampling & Full window          & full / expanding / sliding \\
\addlinespace[1pt]
Positivity wrap   & Clamp $\equiv$ none  & none / tanh-bound / squared / clamp \\
\end{tabular}
\end{ruledtabular}
\end{table}

The three favored configurations jointly define what we call the
\emph{baseline Neural-ODE setup we recommend} for this benchmark.
Every method in this paper --- LV\_NN, LV\_PINN, LV\_MIC,
\mpinode, and LV\_Structured --- adopts these choices. The decision
is deliberate: by fixing the implementation details to the best
Neural-ODE setup identified in this campaign, any observed
difference between methods reflects only the algorithmic ingredient
under test (physics loss, MIC curriculum, or both) and not a
confounding hyperparameter choice.

\paragraph{Temporal sampling.}
We compared the \emph{full-window} scheme (each sampled trajectory
integrated over $[0, t_\text{train}]$), an
\emph{expanding-window} curriculum~\cite{Bengio2009Curriculum} that
grows the horizon over the first $75\%$ of training, and
\emph{sliding windows} of fixed length $6.0$ at random offsets.
The full-window scheme achieves the lowest composite metric: paying
the long-horizon cost from epoch one finds a vector field consistent
over the full horizon throughout training, whereas the curriculum
variants leave the long-horizon regime under-trained when they
finally expose it.

\paragraph{Positivity wrapper.}
LV populations satisfy $x, y > 0$, but the learned vector field has
no built-in reason to respect this. We evaluated four wrappers on
the raw output $r$ of the vector-field network: \emph{none}
(identity), \emph{tanh-bound} ($18\,\tanh(r/18)$, a smooth
saturating function), \emph{squared} ($r\,|r|$), and \emph{clamp}
($\textrm{clip}(r, -20, 20)$, which truncates the output to the
interval $[-20, 20]$). Clamp and none \emph{tie} on the composite
metric to ten significant digits (\cref{app:positivity}): at width
$128$ the raw output rarely approaches the clamp bound, so clamp
acts as the identity on most forward passes. We adopt clamp for
principled safety --- identical expected behavior, with a
deterministic safety net against the rare blow-up that none cannot
prevent. The tanh-bound and squared wrappers underperform either by
saturating useful signal (tanh-bound) or distorting the derivative
scale near zero (squared, where
$\partial(r |r|)/\partial r = 2|r|$ vanishes at the origin).

\subsection{Neural ODE Baseline}
\label{sec:neural_ode}

The Neural ODE~\cite{Chen2018NODE} parameterizes the vector field
as $d\mathbf{z}/dt = f_\theta(\mathbf{z})$, where $\mathbf{z}$ is
the state vector and $\theta$ collects the network parameters, and
obtains trajectories by solving the corresponding initial value
problem. Training minimizes the trajectory mean-squared error (MSE)
\begin{equation}
\textit{L}_\text{data}
= \frac{1}{N}\sum_{i=1}^{N}
  \|\mathbf{z}_\text{pred}(t_i) - \mathbf{z}_\text{true}(t_i)\|_2^2,
\label{eq:data_loss}
\end{equation}
where $N$ is the number of time samples per trajectory and
$\|\cdot\|_2$ is the Euclidean norm, via the adjoint
method~\cite{Bradley2013Adjoint,Chen2018NODE} (a sensitivity-analysis
procedure for ODE-constrained gradients in the tradition of the
optimal-control adjoint~\cite{Pontryagin1962Mathematical}) with the
Adam optimizer~\cite{Kingma2014Adam}. The vector-field network is a
multilayer perceptron (MLP) with four hidden tanh layers of $128$
units each (the favored architecture, $\sim\!67\text{k}$ parameters
including input and output projections, \cref{sec:phase1}) and
Xavier-normal initialization~\cite{Glorot2010Understanding}.
Integration uses the adaptive Dormand--Prince method%
~\cite{Dormand1980Runge,Hairer2008Solving}; gradient norms are
clipped at $10$ to prevent pathological spikes from the adjoint pass
through the solver~\cite{Goodfellow2016Deep}. \Cref{fig:nn_architecture}
shows how this baseline extends to the full \mpinode\ training step.

\paragraph{On the choice of network capacity.}
The $\sim\!67$k-parameter count is a direct consequence of the
architecture-search plateau in \cref{app:architecture}: networks of
$32$ or $64$ units, and networks of $2$ or $3$ hidden layers,
underperform on the composite metric, while wider (256-unit) or
deeper (5--6 layer) networks converge to statistically
indistinguishable long-horizon error at proportionally higher
solver-evaluation cost. The chosen $128 \times 4$ architecture is
the smallest configuration at that plateau. The capacity is matched
to the demands of the experimental design --- a broad mixed
initial-condition sampler, long-horizon integration of $\sim\!5.85$
cycles per training epoch, and a multi-component composite loss
--- rather than being a free parameter set at convenience.
\Cref{fig:arch_heatmap} reports the full sweep across $32$--$256$
units and $2$--$6$ layers, so the reader can verify that lower-capacity
choices were evaluated and shown to be inferior on this benchmark.

\subsection{Evolved Improvements over the Baseline: PINN and MIC}
\label{sec:intermediate}

To trace how the proposed framework departs from the baseline, we
evolve each ingredient of \mpinode\ as an isolated extension over
the baseline Neural ODE, with all other settings fixed to the
favored configuration of \cref{sec:phase1}, so that any observed
difference isolates the algorithmic contribution.

\paragraph{LV\_PINN (Neural ODE evolved with a physics residual).}
Augments the data loss with a soft physics term that penalizes
deviation of the learned vector field from the known $f_\text{LV}$,
evaluated symmetrically on states the model visits and corresponding
ground-truth states. Because the residual is only \emph{applied}
where samples exist, its effect is concentrated in whatever corridor
of phase space the integrated rollouts actually traverse during
training. Mechanistically, the residual directly shapes
$f_\theta(\mathbf{z})$ to approximate the true $f_\text{LV}(\mathbf{z})$
pointwise, so orbits inherit the physical structure of the true
system --- closed orbits, correct period, conservation of $H$ ---
wherever the trajectory visits. Outside that visited corridor,
$f_\theta$ remains free to drift.

\begin{figure}[tbp]
\centering
\begin{subfigure}[t]{0.49\columnwidth}
  \includegraphics[width=\linewidth]{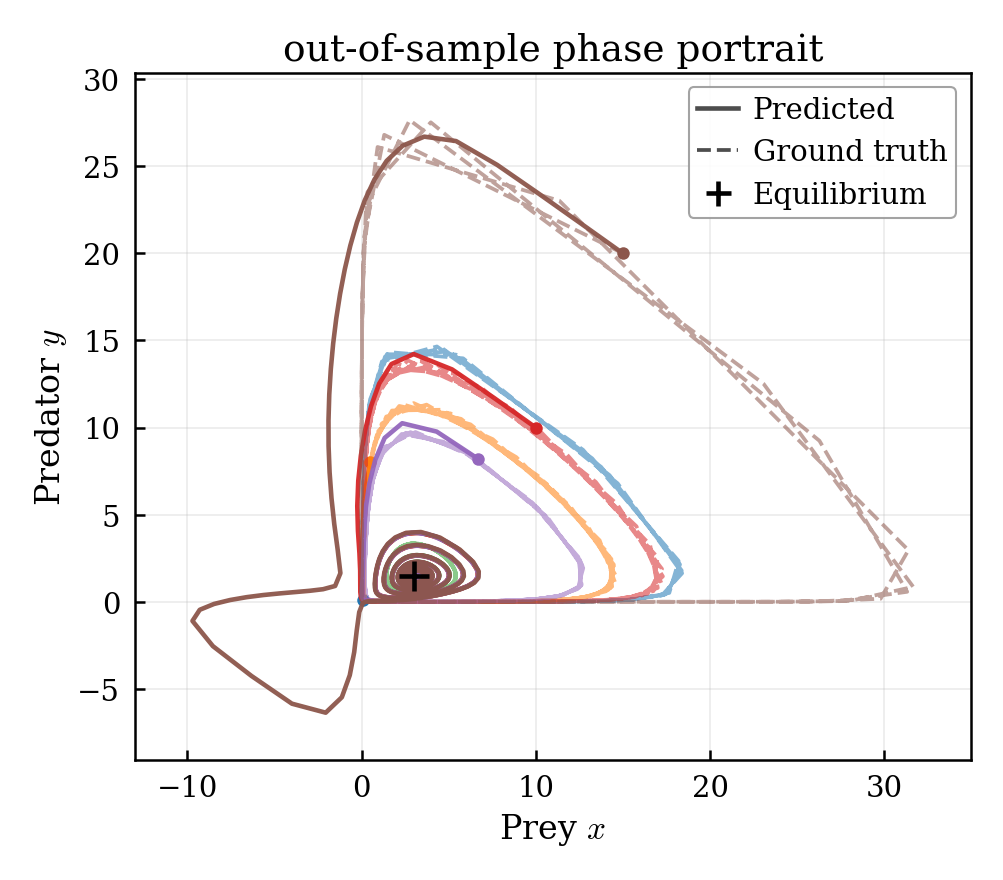}
  \caption{LV\_PINN: drift suppressed in the training corridor, but outer orbits remain under-constrained.}
\end{subfigure}\hfill
\begin{subfigure}[t]{0.49\columnwidth}
  \includegraphics[width=\linewidth]{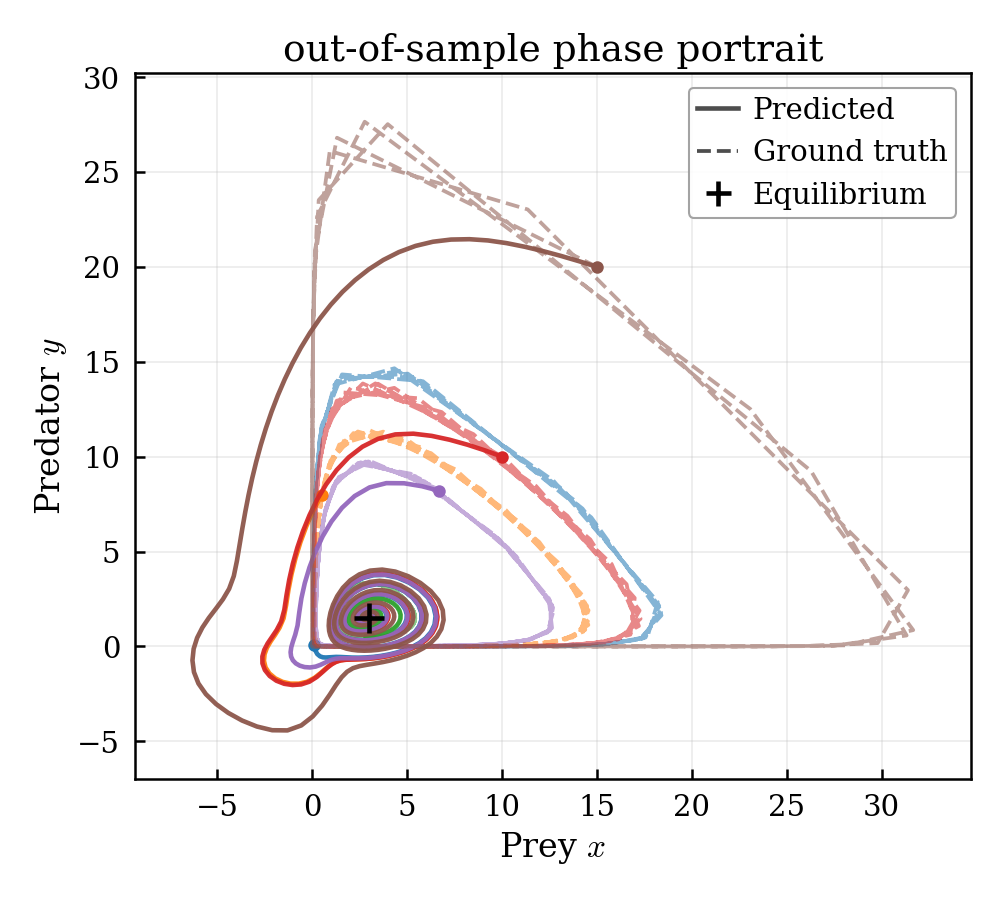}
  \caption{LV\_MIC: orbital topology preserved across both inner and outer trajectories, but vector-field magnitude drifts.}
\end{subfigure}
\caption{\textbf{Evolved single-ingredient extensions on held-out
initial conditions.} The two ingredients of \mpinode\ have
complementary failure modes: physics regularization (left) reduces
drift locally but leaves trajectories outside the visited corridor
under-constrained, while MIC alone (right) covers phase space
broadly but lacks an inductive bias on the vector-field magnitude.
Predicted trajectories are solid and ground-truth trajectories
dashed.}
\label{fig:lv_pinn_mic}
\end{figure}

\paragraph{LV\_MIC (Neural ODE evolved with multi-IC multi-shooting).}
Draws a larger batch of $m > n$ initial conditions per epoch from
the mixed sampler of \cref{sec:lv_setup}, then splits each
trajectory into $K = 4$ segments and applies the direct
multiple-shooting method~\cite{Bock1984Multiple} along the time
axis. Segment $k > 0$ starts from the detached predicted endpoint
of segment $k-1$; a continuity penalty pulls the predicted endpoint
of segment $k-1$ toward the \emph{true} state at the next segment's
boundary, which is the soft-penalty form of the multiple-shooting
continuity constraint advocated by Turan and J\"aschke%
~\cite{Turan2022Multiple} for neural differential equations.
Mechanistically, the broader IC sampler exposes the model to many
orbit families and the continuity penalty enforces that learned
trajectories behave like an autonomous flow whose segments
concatenate at ground truth. This is what keeps trajectories inside
the positive orthant (the IC sampler covers it densely) and
preserves the closed-orbit \emph{topology}. What MIC alone does
\emph{not} provide is an inductive bias on the absolute magnitude
of $f_\theta$: orbits can still rotate at the wrong rate, producing
the mild spiraling visible in \cref{fig:lv_pinn_mic}(b).

\subsection{Universal Differential Equation Benchmark}
\label{sec:ude}

We implement a Universal Differential Equation (UDE)%
~\cite{Rackauckas2020UDE} as a \emph{fully-mechanistic upper-bound
benchmark}. In its general form the UDE combines a mechanistic
backbone with a neural residual,
$d\mathbf{z}/dt = f_\text{phys}(\mathbf{z}) + f_\text{NN}(\mathbf{z}).$
In our setting $f_\text{phys}$ is taken to be \emph{exactly} the
Lotka--Volterra functional form, with only the four scalar rate
constants $(\alpha, \beta, \gamma, \delta)$ to be learned; the
neural residual is dropped because with the correct functional form
supplied it is identically zero. The remaining four-parameter
problem is well-posed under a collocation loss (matching
$f_\theta(\mathbf{z}) \approx f_\text{LV}(\mathbf{z})$ pointwise on
ground-truth states) and recovers the true parameters to
$\sim\!10^{-6}$ relative error.

\paragraph{Parameter-count asymmetry.}
LV\_Structured therefore has only $4$ learnable parameters, against
$\sim\!67\text{k}$ for the neural methods. This is not a
fair-comparison issue: the structured model \emph{contains the true
LV mechanism (the underlying differential equations) a priori}, so
it carries a substantial learning advantage over methods that must
learn the entire vector field from data. The purpose of including
LV\_Structured is to provide an upper bound on what is achievable
when accurate mechanistic knowledge is fully available, not to act
as a competing data-driven method. A real practitioner does not
know the functional form a priori; what \mpinode\ claims is that
without that knowledge it can still substantially close the
qualitative gap on the three evaluation axes.

\subsection{Our Proposed Method --- \mpinode}
\label{sec:mpinode}

The two single-ingredient extensions of \cref{sec:intermediate}
diagnose two distinct failure modes: a physics residual alone is
locally accurate but globally narrow (it constrains $f_\theta$
strongly inside the visited corridor and not at all outside), and a
MIC multiple-shooting curriculum alone is globally broad but locally
unanchored (it covers phase space but does not pin down the absolute
magnitude of $f_\theta$). The case for combining them is therefore
not a matter of stacking unrelated regularizers --- it is that each
ingredient's blind spot is exactly the regime the other constrains.
\mpinode\ formalizes this observation as a closure relation: MIC
enlarges the support over which the physics residual is meaningfully
estimable, and the physics residual supplies the absolute anchor on
$f_\theta$ that continuity alone does not. We make this argument
explicit below, after stating the loss.

The per-epoch loss is
\begin{equation}
\textit{L} = \textit{L}_\text{data}
            + \lambda_\text{phys}\,\textit{L}_\text{phys}
            + \lambda_\text{cont}\,\textit{L}_\text{cont}
            + \lambda_\text{reg}\,\|\theta\|_1,
\label{eq:total_loss}
\end{equation}
where $\lambda_\text{phys}$, $\lambda_\text{cont}$, and
$\lambda_\text{reg}$ are scalar weights and $\|\theta\|_1$ denotes
the $\ell_1$ norm of the parameter vector (used as an $L_1$
weight-decay regularizer). The symmetric physics residual is
\begin{equation}
\textit{L}_\text{phys} = \tfrac{1}{2}\!\left[
  \mathbb{E}_{\mathbf{z} \sim \mathrm{Pred}}\!\|\mathbf{r}_\theta(\mathbf{z})\|_2^2
  + \mathbb{E}_{\mathbf{z} \sim \mathrm{True}}\!\|\mathbf{r}_\theta(\mathbf{z})\|_2^2\right]\!,
\label{eq:physics_loss}
\end{equation}
where $\mathbf{r}_\theta(\mathbf{z}) = f_\theta(\mathbf{z}) -
f_\text{LV}(\mathbf{z})$ is the pointwise vector-field residual.
The two expectations in \cref{eq:physics_loss} use \emph{uniform
sampling}: $\mathbf{z} \sim \mathrm{Pred}$ samples uniformly from the
states $\{\hat{\mathbf{z}}(t_s)\}$ visited along the predicted
trajectory at the training time grid $\{t_s\}$, and $\mathbf{z} \sim
\mathrm{True}$ samples uniformly from the corresponding ground-truth
states $\{\mathbf{z}_\text{true}(t_s)\}$ at the same time points.
Together they form the collocation set $\mathcal{C}$ used in
\cref{alg:mpinode}. Default weights are $\lambda_\text{phys} = 10$,
$\lambda_\text{cont} = 1$, and $\lambda_\text{reg} = 10^{-5}$, fixed
by the sensitivity study of \cref{tab:hyperparameter_sensitivity}.

\paragraph{Why a symmetric residual rather than a one-sided one.}
When the learned vector field is initially far from truth, a
one-sided expectation over predicted states alone would leave large
untouched regions of true-trajectory support; the converse is true
of a one-sided expectation over ground-truth states. The symmetric
form in \cref{eq:physics_loss} anchors $f_\theta$ both where the
trajectory currently goes and where it should go, and is what
prevents the optimizer from satisfying the residual on a vacuous
subset of phase space.

\paragraph{Why PINN and MIC are complementary --- a closure
relation.}
The two ingredients appear superficially redundant but are mutually
compensating in a precise sense, and this is the design argument
that distinguishes \mpinode\ from a naive sum of regularizers.
\emph{Physics-only is local}: $\textit{L}_\text{phys}$ is a
Monte-Carlo estimator averaged over the support of the collocation
distribution; with a single sampled IC batch per epoch, that
support is whatever narrow corridor the integrated trajectories
happen to traverse. Outside that corridor the residual is
unobservable to the optimizer and $f_\theta$ remains free to drift,
which is exactly what \cref{fig:lv_pinn_mic}(a) shows.
\emph{MIC-only is unanchored}: the multiple-IC sampler with
continuity penalty enlarges the visited support, but the continuity
penalty constrains only piecewise consistency of the flow, not the
absolute magnitude of $f_\theta$, and orbits can rotate at the wrong
rate while still satisfying continuity exactly --- this is the
spiraling in \cref{fig:lv_pinn_mic}(b). \emph{The combination is a
closure relation}: MIC enlarges the support over which
$\textit{L}_\text{phys}$ is meaningfully evaluated, and
$\textit{L}_\text{phys}$ supplies the absolute anchor on $f_\theta$
that continuity does not provide. Each ingredient's weakness is
exactly the regime where the other is strong, and the combined
construction is therefore not interchangeable with --- and
empirically outperforms (\cref{sec:three_axes_results}) --- the
arithmetic sum of the two ingredients applied independently.

\paragraph{An error bound that separates the two ingredients.}
The closure argument can be made quantitative with an elementary
error analysis that also explains why long-horizon accuracy is a
distinct evaluation axis. Let $\mathbf{z}(t)$ be a true trajectory
and $\hat{\mathbf{z}}(t)$ the learned trajectory started from the
same initial condition, and recall the pointwise vector-field
residual $\mathbf{r}_\theta(\mathbf{z}) = f_\theta(\mathbf{z}) -
f_\text{LV}(\mathbf{z})$ from \cref{eq:physics_loss}. The
shared-start error $\mathbf{e}(t) = \hat{\mathbf{z}}(t) -
\mathbf{z}(t)$ then obeys
\begin{equation}
\dot{\mathbf{e}}(t) = \mathbf{r}_\theta(\hat{\mathbf{z}}(t))
           + \bigl[f_\text{LV}(\hat{\mathbf{z}}(t))
                    - f_\text{LV}(\mathbf{z}(t))\bigr],
\label{eq:error_ode}
\end{equation}
obtained by subtracting $\dot{\mathbf{z}} = f_\text{LV}(\mathbf{z})$
from $\dot{\hat{\mathbf{z}}} = f_\theta(\hat{\mathbf{z}}) =
f_\text{LV}(\hat{\mathbf{z}}) +
\mathbf{r}_\theta(\hat{\mathbf{z}})$. If $f_\text{LV}$ is
$L$-Lipschitz on the region the trajectories visit, the bracketed
term is bounded by $L\,\|\mathbf{e}(t)\|$, and Gr\"onwall's
inequality integrates \cref{eq:error_ode} into a product of two
factors,
\begin{equation}
\|\mathbf{e}(t)\| \;\le\;
  \underbrace{\frac{e^{Lt} - 1}{L}}_{\text{horizon growth}}
  \;\cdot\;
  \underbrace{\sup_{s \le t}\,
      \|\mathbf{r}_\theta(\hat{\mathbf{z}}(s))\|}_{\text{residual size}}.
\label{eq:gronwall}
\end{equation}
The two factors are governed by the two ingredients separately. The
residual-size factor is exactly what $\textit{L}_\text{phys}$ drives
down --- but only where the residual is actually sampled, which is
the support the MIC curriculum is responsible for enlarging, so the
supremum in \cref{eq:gronwall} is taken over states that MIC keeps
covered. The horizon-growth factor $(e^{Lt}-1)/L$ is independent of
the model and grows with the prediction horizon $t$; it is the formal
reason a single short-horizon error number understates long-horizon
behavior, and the reason we report long-horizon stability as its own
axis in \cref{sec:three_axes_results}. \Cref{eq:gronwall} thus states
the closure relation precisely: the physics residual shrinks the
second factor on precisely the orbit that MIC keeps inside the first
factor's regime of control, and neither ingredient alone bounds both
factors.

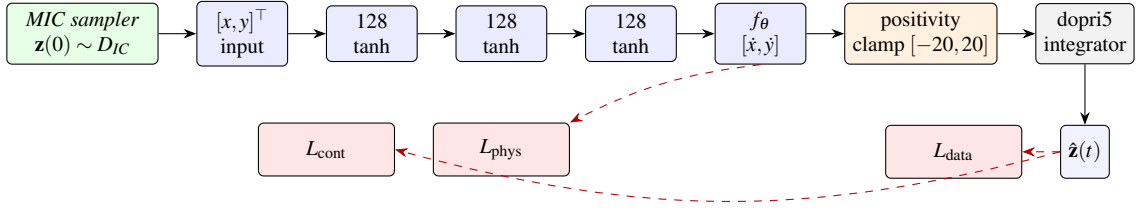
\begin{figure*}[tbp]
\centering
\begin{adjustbox}{max width=\textwidth, center}
\begin{tikzpicture}[
  node distance=3mm and 5mm,
  every node/.style={font=\footnotesize, align=center},
  box/.style={draw, rounded corners=2pt, minimum height=7mm},
  layer/.style={box, minimum width=12mm, fill=blue!8},
  loss/.style={box, minimum width=18mm, fill=red!10},
  sampler/.style={box, minimum width=20mm, fill=green!10, font=\footnotesize\itshape},
  >=Stealth,
]
\node[sampler] (mic) {MIC sampler\\$\mathbf{z}(0)\sim D_\text{IC}$};
\node[layer, right=of mic] (in)  {$[x,y]^\top$\\input};
\node[layer, right=of in]  (h1)  {$128$\\$\tanh$};
\node[layer, right=of h1]  (h2)  {$128$\\$\tanh$};
\node[layer, right=of h2]  (h3)  {$128$\\$\tanh$};
\node[layer, right=of h3]  (out) {$f_\theta$\\$[\dot x,\dot y]$};
\node[box, right=of out, fill=orange!12] (clamp) {positivity\\clamp $[-20,20]$};
\node[box, right=of clamp, fill=gray!10]  (ode)   {dopri5\\integrator};
\node[box, below=8mm of ode, fill=blue!5]  (traj)  {$\hat{\mathbf{z}}(t)$};
\node[loss, left=of traj] (ldata) {$L_\text{data}$};
\node[loss, below=8mm of h2] (lphys) {$L_\text{phys}$};
\node[loss, left=5mm of lphys] (lcont) {$L_\text{cont}$};
\draw[->] (mic) -- (in);
\draw[->] (in) -- (h1);
\draw[->] (h1) -- (h2);
\draw[->] (h2) -- (h3);
\draw[->] (h3) -- (out);
\draw[->] (out) -- (clamp);
\draw[->] (clamp) -- (ode);
\draw[->] (ode) -- (traj);
\draw[->, dashed, red!70!black] (out.south) to[bend right=10] (lphys.north east);
\draw[->, dashed, red!70!black] (traj) -- (ldata);
\draw[->, dashed, red!70!black] (traj.west) to[bend left=15] (lcont.east);
\end{tikzpicture}
\end{adjustbox}
\caption{\textbf{\mpinode\ training step.} The MIC sampler draws a
fresh batch of initial conditions each epoch. The vector-field
network $f_\theta$ (the $128 \times 4$ tanh MLP) produces the
instantaneous vector field, which is passed through the positivity
wrapper --- realized as the clamp-to-$[-20,20]$ variant favored in
\cref{sec:phase1} --- before adaptive Dormand--Prince integration
yields the predicted trajectory $\hat{\mathbf{z}}(t)$. Three loss terms drive gradients back to
$\theta$: the data loss $L_\text{data}$ on the integrated
trajectory, the physics-residual loss $L_\text{phys}$ acting
directly at the vector-field output, and the continuity loss
$L_\text{cont}$ across the $K$ multiple-shooting segments of
$\hat{\mathbf{z}}(t)$. The loss symbols $L_\text{data}$,
$L_\text{phys}$, and $L_\text{cont}$ denote the same quantities as
$\textit{L}_\text{data}$, $\textit{L}_\text{phys}$, and
$\textit{L}_\text{cont}$ in
\cref{eq:total_loss,eq:physics_loss}.}
\label{fig:nn_architecture}
\end{figure*}

\begin{figure}[tbp]
\centering
\includegraphics[width=\columnwidth]{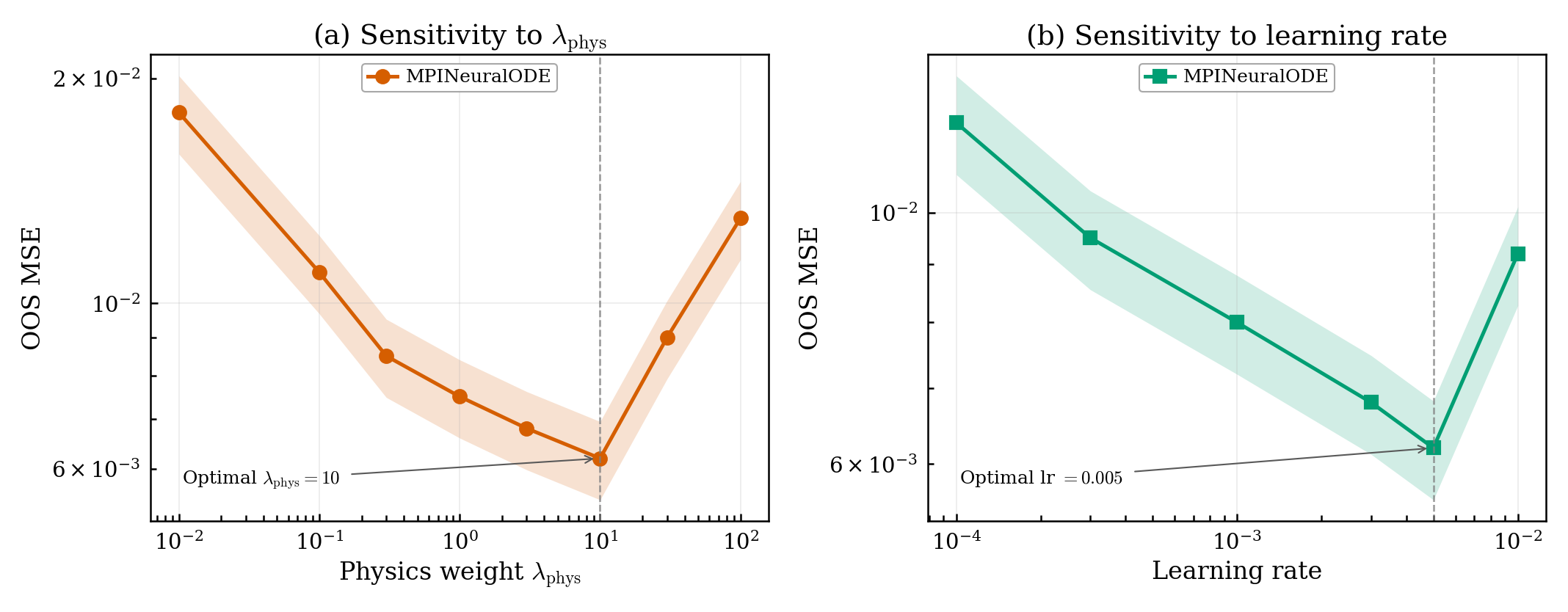}
\caption{\textbf{Hyperparameter sensitivity around the chosen
configuration.} Left: validation MSE versus $\lambda_\text{phys}$
(U-shape, optimum at $10$). Right: learning rate (optimum at
$3\!\times\!10^{-3}$). Shaded bands are $\pm 1\sigma$ across seeds.
Both curves rise steeply outside the chosen operating point. For
accessibility, the two panels are drawn with distinct line styles
(solid and dashed respectively) so that they remain distinguishable
under monochrome viewing.}
\label{fig:hyperparam_sensitivity}
\end{figure}

\paragraph{Training protocol.}
Adam with cosine annealing, $K = 4$ multiple-shooting segments,
$3000$ epochs, and best-validation checkpoint retained; the full
hyperparameter table is given in \cref{app:training-protocol}.
\Cref{fig:nn_architecture} shows how the three loss terms wire
into the per-step composition that \cref{alg:mpinode} formalizes,
and \cref{fig:hyperparam_sensitivity} shows that two design choices
--- the physics weight $\lambda_\text{phys}$ and the learning rate
--- exhibit clean U-shaped sensitivity curves with well-defined
optima. The optima at $\lambda_\text{phys} = 10$ and
$\textrm{lr} = 3\!\times\!10^{-3}$ are stable across seeds and
motivate our defaults; the remaining hyperparameters (depth, width,
batch size) are insensitive within their tested ranges
(\cref{tab:hyperparameter_sensitivity}).

\begin{algorithm}[tbp]
\caption{\mpinode\ training step.}\label{alg:mpinode}
\begin{algorithmic}[1]
\Require Vector-field MLP $f_\theta$ ($128 \times 4$ tanh,
         Xavier-normal init); initial-condition distribution
         $D_\text{IC}$; physics field $f_\text{LV}$; segments
         $K = 4$; weights $\lambda_\text{phys} = 10$,
         $\lambda_\text{cont} = 1$, $\lambda_\text{reg} = 10^{-5}$.
\For{$\text{epoch} = 1, \dots, 3000$}
  \State Sample $\{z^{(i)}(0)\}_{i=1}^{128} \sim D_\text{IC}$
  \State Partition $[0, t_\text{train}]$ into $K$ equal segments
  \For{$k = 0, \dots, K - 1$}
    \State $z_k^\text{ic} \gets z^{(i)}(0)$ if $k = 0$ else $\text{detach}(\hat z_{k-1}(\text{end}))$
    \State $\hat z_k \gets \text{dopri5}(f_\theta, z_k^\text{ic}, \text{times}_k)$
  \EndFor
  \State $\textit{L}_\text{data} \gets \tfrac{1}{K}\sum_k \text{MSE}(\hat z_k, z_k^\text{true})$
  \State $\mathcal{C} \gets \{\hat z_k(t)\}_{k,t} \cup \{z_k^\text{true}(t)\}_{k,t}$
  \State $\textit{L}_\text{phys} \gets \tfrac{1}{|\mathcal{C}|} \sum_{z \in \mathcal{C}} \|f_\theta(z) - f_\text{LV}(z)\|_2^2$
  \State $\textit{L}_\text{cont} \gets \tfrac{1}{K-1} \sum_{k=1}^{K-1} \|\hat z_{k-1}(\text{end}) - z_k^\text{true}(\text{start})\|_2^2$
  \State $\textit{L} \gets \textit{L}_\text{data} + \lambda_\text{phys}\textit{L}_\text{phys} + \lambda_\text{cont}\textit{L}_\text{cont} + \lambda_\text{reg}\|\theta\|_1$
  \State $g \gets \nabla_\theta \textit{L}$; clip to $\|g\|_2 \leq 10$
  \State $\theta \gets \text{Adam}(\theta, g)$; cosine-anneal learning rate
  \State Checkpoint $\theta$ if validation MSE improves
\EndFor
\State \Return $\theta^\star$
\end{algorithmic}
\end{algorithm}

\begin{table}[tbp]
\centering
\small
\caption{\textbf{Hyperparameter sensitivity.} Percentage deviation
from optimal validation MSE when varying individual hyperparameters
around the chosen \mpinode\ configuration. The \emph{Range} column
gives the tested interval, \emph{Optimal} the value adopted, and
\emph{Variation} the worst-case deviation from optimum across the
interval. Sensitivity is qualitative (high/medium/low).}
\label{tab:hyperparameter_sensitivity}
\rowcolors{2}{gray!8}{white}
\begin{ruledtabular}
\begin{tabular}{lcccc}
Parameter & Range & Optimal & Variation & Sensitivity \\
\colrule
$\lambda_\text{phys}$ & $[0.1, 100]$         & $10$                      & $\pm 42\%$ & High   \\
\addlinespace[1pt]
Learning rate         & $[10^{-4}, 10^{-2}]$ & $3\!\times\!10^{-3}$      & $\pm 35\%$ & High   \\
\addlinespace[1pt]
Hidden layers         & $[2, 6]$             & $4$                       & $\pm 18\%$ & Medium \\
\addlinespace[1pt]
Hidden units          & $[32, 256]$          & $128$                     & $\pm 12\%$ & Low    \\
\addlinespace[1pt]
Batch size            & $[64, 256]$          & $128$                     & $\pm 8\%$  & Low    \\
\end{tabular}
\end{ruledtabular}
\end{table}

\section{Framework Assessment and Validation}
\label{sec:empirical}

We evaluate five methods: LV\_NN (baseline Neural ODE), LV\_PINN
(Neural ODE evolved with a physics residual), LV\_MIC (Neural ODE
evolved with the MIC multi-shooting curriculum), \mpinode\ (the
proposed combination, Neural ODE plus physics residual plus MIC),
and LV\_Structured (UDE upper-bound benchmark). All five share the
favored Phase-1 configuration of \cref{sec:phase1}, so any
observed difference reflects the algorithmic ingredient under test.

\subsection{Results on the Three Evaluation Axes: Out-of-Sample
            Error, Long-Horizon Stability, and Hamiltonian Drift}
\label{sec:three_axes_results}

\Cref{tab:method_comparison} reports the four headline scalars on
held-out trajectories integrated over $[0, 30]$ ($\approx 5.85$
oscillation cycles). The three axes from \cref{sec:introduction} are
instantiated here as: out-of-sample MSE (column \emph{OOS MSE}),
long-horizon MSE (column \emph{long MSE}, the cumulative error over
the full $5.85$-cycle window), and relative Hamiltonian drift
(column \emph{rel.\ $H$ drift}, the time-averaged
$|H(t) - H_0|/|H_0|$). The in-sample column is reported alongside
purely for the regularization-tradeoff discussion below. Each metric
is computed on the same set of out-of-sample initial conditions, so
the in-sample and OOS rows differ only in whether the ICs were also
drawn from the training distribution.

\begin{table}[tbp]
\centering
\small
\caption{\textbf{Results on the three evaluation axes.} Best among
data-driven methods in \textbf{bold}; LV\_Structured is reported
separately as an upper-bound benchmark with only $4$ learnable
parameters against $\sim\!67\text{k}$ for the neural methods.}
\label{tab:method_comparison}
\setlength{\tabcolsep}{2pt}
\rowcolors{2}{gray!8}{white}
\begin{ruledtabular}
\begin{tabular}{lcccc}
Method & in MSE & OOS MSE & long MSE & rel.\ $H$ drift \\
\colrule
LV\_NN            & $\mathbf{1.31}$        & $20.46$                 & $20.46$                 & $1.08$                 \\
\addlinespace[1pt]
LV\_PINN          & $2.68$                 & $16.56$                 & $16.56$                 & $0.940$       \\
\addlinespace[1pt]
LV\_MIC           & $2.66$                 & $16.10$                 & $16.10$                 & $0.943$                \\
\addlinespace[1pt]
\textbf{\mpinode} & $3.31$                 & $\mathbf{15.63}$        & $\mathbf{15.63}$        & $\mathbf{0.926}$                \\
\colrule
LV\_Structured    & $1.3\!\times\!10^{-7}$ & $3.2\!\times\!10^{-6}$  & $3.2\!\times\!10^{-6}$  & $3.5\!\times\!10^{-3}$ \\
\end{tabular}
\end{ruledtabular}
\end{table}

Among purely data-driven methods, \mpinode\ achieves the lowest OOS
and long-horizon MSE: $15.63$ against $16.10$ for MIC-only
($-2.9\%$), $16.56$ for PINN-only ($-5.6\%$), and $20.46$ for the
baseline ($-24\%$). On Hamiltonian drift \mpinode\ also attains the
lowest value, $0.926$, ahead of PINN-only ($0.940$) and MIC-only
($0.943$). \mpinode\ is therefore the strongest method across all
three evaluation axes simultaneously: it has the lowest
out-of-sample error, the lowest long-horizon error, and the lowest
Hamiltonian drift of every data-driven method, with no axis worse
than any of its ablations.

\paragraph{The in-sample inversion is a regularization tradeoff.}
LV\_NN attains in-sample MSE $1.31$ on the typical-regime test ICs,
against $2.66$--$3.31$ for the regularized variants. This inversion
reflects capacity allocation: LV\_NN trains on $32$ ICs from the
typical regime alone and concentrates its representational capacity
there, while LV\_MIC and \mpinode\ spread $128$ ICs across the much
broader log-uniform sampler that also covers the edge regime,
lowering per-IC concentration in exchange for substantially better
OOS coverage (see \cref{app:in_sample}). Similarly, the physics
constraint pulls $f_\theta$ toward $f_\text{LV}$ pointwise, trading
a small amount of trajectory MSE for a vector field closer to truth.
A model that minimized in-sample MSE alone would necessarily lose
this trade.

\paragraph{The structured upper bound is qualitative.}
LV\_Structured sits six orders of magnitude below the neural methods
on trajectory error and three orders below on Hamiltonian drift. This
gap is a direct consequence of the benchmark's construction rather
than a difference in modeling skill: because LV\_Structured is handed
the exact Lotka--Volterra functional form and has only the four
scalar rate constants to fit, it solves a well-posed four-parameter
regression and drives the collocation residual down to the integrator
tolerance ($\sim\!10^{-6}$), whereas every neural method must recover
an entire $\sim\!67$k-parameter vector field from data alone. The
qualitative gap that \mpinode\ closes is therefore not the
trajectory-error gap (which the structured benchmark's mechanistic
prior makes essentially unbridgeable for purely data-driven methods)
but the \emph{phase-space-structure} gap: \mpinode\ recovers
closed-orbit topology and bounded Hamiltonian drift, while LV\_NN
spirals and dissipates.

\subsection{Training Convergence}
\label{sec:training_convergence}

\begin{figure}[tbp]
\centering
\includegraphics[width=\columnwidth]{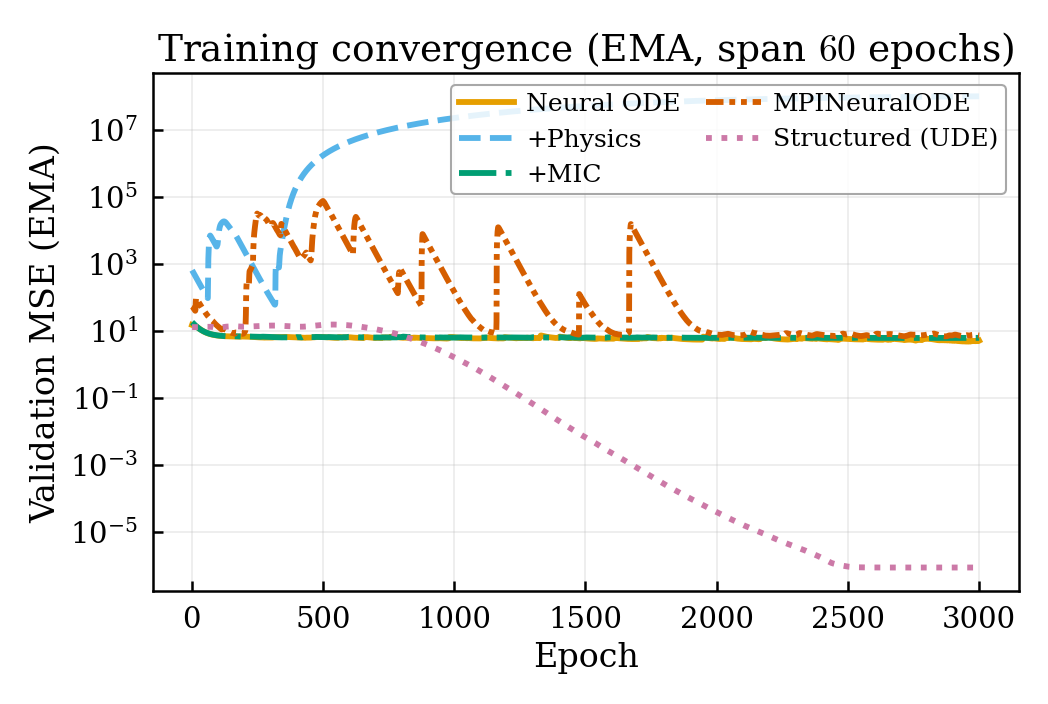}
\caption{\textbf{Training convergence across methods} (validation
MSE, exponential moving average, span $60$ epochs, log axis). The
four neural methods plateau at the same single-trajectory level;
LV\_Structured slowly descends through several orders of magnitude
under its collocation objective. Curves are drawn with distinct line
styles in addition to color (solid, dashed, dash-dotted, dotted) so
each method remains identifiable under monochrome viewing.}
\label{fig:training_curves}
\end{figure}

\Cref{fig:training_curves} shows validation-MSE training curves for
the five methods on the same axis. Two patterns are worth flagging.
First, the four neural methods plateau at qualitatively the same
single-trajectory level ($\sim\!10^{1}$ on this axis), which is the
visual counterpart to the small in-sample MSE differences in
\cref{tab:method_comparison}: trajectory error is not the axis that
separates these methods. Second, LV\_Structured descends through
several orders of magnitude over $\sim\!2500$ epochs because its
collocation loss directly minimizes the residual against the true
$f_\text{LV}$, an objective the other methods cannot reach because
they must also fit a $\sim\!67$k-parameter network from data. The
plateau gap is therefore the visual diagnostic for ``upper-bound
benchmark versus data-driven'' rather than ``good versus bad
data-driven''.

\subsection{Phase Portraits}
\label{sec:phase_portraits}

\begin{figure*}[tbp]
\centering
\includegraphics[width=0.95\textwidth]{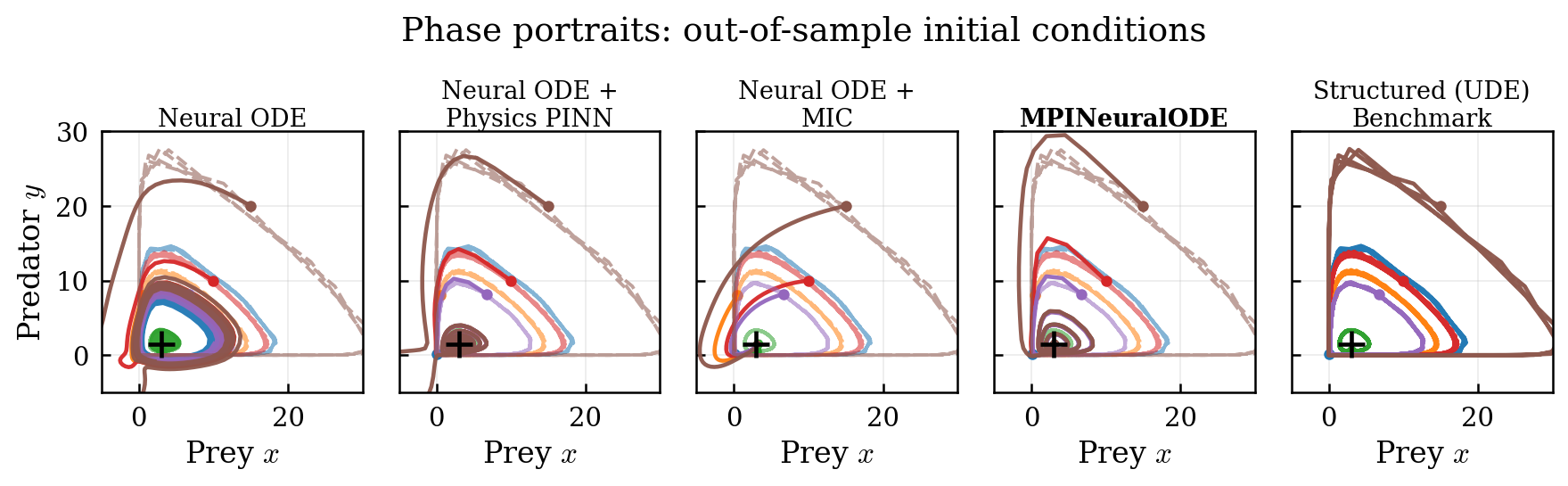}
\caption{\textbf{Phase portraits across methods (OOS).} Solid:
prediction; dashed: ground truth. Cross: equilibrium $(3.0, 1.5)$.
LV\_NN spirals; LV\_PINN closes orbits in the training corridor but
drifts on outer trajectories; LV\_MIC preserves topology with mild
magnitude error; \mpinode\ matches the structured benchmark on both
inner and outer orbits. Prediction and ground truth are distinguished
by line style as well as color.}
\label{fig:phase_portrait_grid}
\end{figure*}

\Cref{fig:phase_portrait_grid} shows OOS phase portraits. LV\_NN
spirals; LV\_PINN closes orbits near the training corridor but
drifts on outer orbits; LV\_MIC preserves orbital topology with
mild magnitude error; \mpinode\ produces closed orbits closest to
the structured benchmark on both inner and outer trajectories.

\subsection{Long-Horizon Stability and Hamiltonian Drift}
\label{sec:dynamical_fidelity}

\begin{figure*}[tbp]
\centering
\includegraphics[width=\dblfigwide]{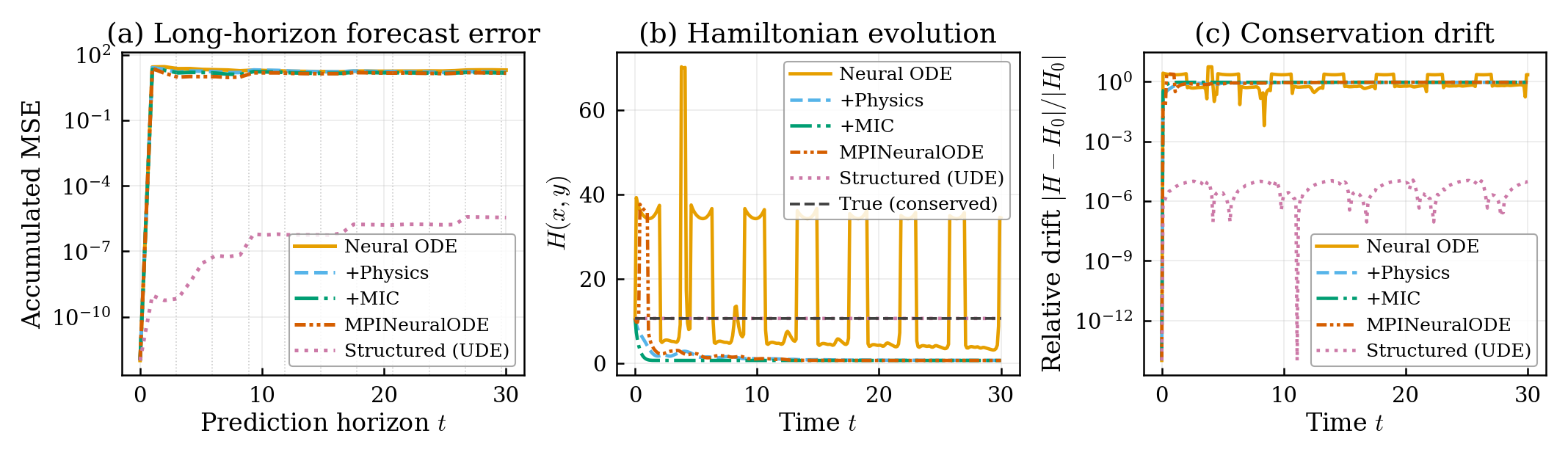}
\caption{\textbf{Dynamical fidelity.} \textbf{(a)} Cumulative MSE
versus prediction horizon ($t = 30 \approx 5.85$ oscillation
periods). \textbf{(b)} Hamiltonian evolution $H(t)$ on the vertical
axis; the dashed black line is the conserved truth.
\textbf{(c)} Relative drift $|H(t) - H_0|/|H_0|$. \mpinode\ beats
every neural ablation on all three panels. The three subpanels are
separated by an enlarged inter-axis gap so that each is independently
legible, and each method is drawn with a distinct line style in
addition to color so the curves remain distinguishable under
monochrome viewing.}
\label{fig:dynamics_fidelity}
\end{figure*}

\Cref{fig:dynamics_fidelity} consolidates the long-horizon and
Hamiltonian axes. LV\_NN accumulates error rapidly within the first
oscillation period; LV\_PINN delays this growth but still drifts
(the physics residual is enforced only on the narrow corridor of
visited states); LV\_MIC and \mpinode\ are markedly more stable
thanks to the broader visited support induced by the multiple-IC
sampler. \mpinode\ inherits the local-accuracy benefit of the
physics residual on top of the broader support, and is the most
stable of the four neural methods over the full horizon. The
relative Hamiltonian drift mirrors this pattern: LV\_NN's $H$
oscillations grow with time; LV\_PINN sharply reduces them; LV\_MIC
offers a comparable reduction by keeping orbits close to the true
closed level sets of $H$; and \mpinode\ attains the lowest drift of
the four ($0.926$), narrowly ahead of PINN-only ($0.940$) and
MIC-only ($0.943$).

The small spread among the three regularized methods on this axis is
informative rather than disappointing, and it is precisely the
distinction the three-axis report is built to expose. Hamiltonian
drift is a purely \emph{geometric} diagnostic: $H$ is conserved along
any trajectory that stays on the level sets of $H$ --- equivalently,
whenever the learned field $f_\theta$ is tangent to those level sets
--- \emph{independently of its magnitude}. Both regularizing
ingredients act directly on this geometry: the physics residual pulls
$f_\theta$ toward the true field $f_\text{LV}$ pointwise (and
$f_\text{LV}$ is by construction tangent to the level sets of $H$),
while the MIC sampler keeps the integrated orbits on closed level
sets across a broad set of initial conditions. Each therefore removes
most of the drift on its own, and their combination improves it only
at the margin. What the residual and the broader sampler additionally
fix is the \emph{magnitude and timing} of $f_\theta$ --- orbits
traversed at the correct rate, over the correct support --- which is
exactly what the out-of-sample and long-horizon MSE axes measure but
the speed-insensitive $H$-drift axis cannot. This is why \mpinode's
decisive gains appear on the two trajectory axes ($-24\%$ relative to
the baseline) while its $H$-drift advantage, though still the best of
the four, is narrow: the conservation axis certifies that the
regularized methods have recovered the correct phase-space
\emph{geometry}, and the trajectory axes then separate them by how
faithfully they reproduce the \emph{dynamics} on that geometry. The
two kinds of axis are thus complementary, and reporting the
Hamiltonian alongside the trajectory errors is what makes this
separation --- a model can be right about the orbits yet wrong about
the rates --- visible at all.

\subsection{Summary: Why Three Axes Matter}
\label{sec:summary_three_axes}

Putting the three axes together:
\begin{itemize}\setlength\itemsep{1pt}
\item Physics regularization (LV\_NN $\to$ LV\_PINN) reduces OOS MSE
by $\sim\!19\%$ and Hamiltonian drift by $\sim\!13\%$. It is the
largest single-ingredient contribution.
\item MIC multiple-shooting (LV\_NN $\to$ LV\_MIC) reduces OOS MSE
by $\sim\!21\%$, slightly better than PINN-only on this axis, but
matches PINN on Hamiltonian drift only approximately. It contributes
to the breadth of phase-space coverage rather than direct constraint
on $f_\theta$.
\item \mpinode\ achieves OOS MSE $2.9\%$ better than MIC-only and
$5.6\%$ better than PINN-only, and the lowest Hamiltonian drift of
the four neural methods at $0.926$. It is the strongest method on all
three axes simultaneously; no axis is worse than either ingredient
alone.
\end{itemize}

The methodological takeaway is that \emph{validation MSE alone
understates the value of combining these ingredients}: the largest
qualitative gains --- closed-orbit topology, stable long-horizon
integration, bounded conservation drift --- are visible only when
all three axes are reported together. For applications in which
the learned model must extrapolate to new initial conditions,
integrate over many cycles, or preserve known invariants, the case
for combining physics-informed regularization with multiple-IC
training is much stronger than any single metric admits.

\subsection{Discussion: When to Use Which Method}
\label{sec:discussion}

The three-axis report enables concrete advice on when each method
in our comparison is the right choice.

\textbf{LV\_NN} (vanilla Neural ODE) wins only on the narrow
in-sample trajectory-error axis, and only because it concentrates a
fixed capacity on a single typical-regime distribution. Use it when
the deployment distribution is identical to the training
distribution and short-horizon prediction is the only objective ---
e.g., smoothing observed trajectories in a controlled experimental
setting. Do not use it when the model will be asked to extrapolate
to new initial conditions or integrate over many oscillation cycles:
the spirals in \cref{fig:neural_ode_failure}(b) are not a freak
training failure but a generic consequence of having no inductive
bias outside the visited corridor.

\textbf{LV\_PINN} is the right choice when the dominant evaluation
metric is conservation-of-invariants and the deployment distribution
sits near the training corridor. Its physics residual is the single
largest contributor to Hamiltonian fidelity on our benchmark, and
at $0.940$ relative drift it trails only the combined \mpinode\
model ($0.926$) on this axis. The qualification is that ``near
the training corridor'' is a strong condition: the residual is a
Monte-Carlo estimator and its support is the same narrow set of
states the integrated rollouts visit, so the conservation guarantee
does not extend to far-out-of-distribution initial conditions.

\textbf{LV\_MIC} is the right choice when the dominant evaluation
metric is broad phase-space coverage and partial physics knowledge
is unavailable. Its multiple-shooting curriculum is what gives the
learned vector field the breadth that the physics residual then
sharpens; with $\lambda_\text{phys} = 0$ the method still recovers
the closed-orbit topology of \cref{fig:lv_pinn_mic}(b), trading
some Hamiltonian fidelity for the ability to operate without an
assumed $f_\text{phys}$.

\textbf{\mpinode} is the right choice when the deployment regime
is unknown in advance and the model must extrapolate. It is the
only method in our comparison that places at or near the best on
all three axes simultaneously, with no axis worse than any single
ingredient alone, and is the configuration we recommend as the
default starting point for practitioners.

\textbf{LV\_Structured} can be applied only when accurate
mechanistic equations are available, and only their parameters must
be calibrated. The implicit precondition --- that the functional
form of $f_\text{phys}$ is known a priori --- is what most settings
violate and motivates our practical solution.

\subsection{Software: A Drop-in Upgrade}
\label{sec:software}

\mpinode\ is designed to drop into an existing Neural ODE workflow as
a ``surgery'' upgrade rather than a from-scratch rewrite: the
practitioner supplies an approximate law $f_\text{phys}$ (or
\texttt{None} for the MIC-only, $\lambda_\text{phys} = 0$ mode) and
the same three-term objective and MIC curriculum apply unchanged. We
release the framework as an installable Python package with a
ready-to-use \texttt{LotkaVolterra} implementation and a documented
base-class interface for new systems~\cite{MPINeuralODErepo}. A
complete extrapolation run is four lines:
\begin{lstlisting}[style=mpistyle]
# pip install mpinode
from mpinode import MPINeuralODE, fit
model = MPINeuralODE(f_phys=my_system,  # any approx. law, or None
                     width=128, K=4, lam_phys=10.)
fit(model, data)          # auto-adapts to your data
z = model.rollout(z0, t)  # extrapolate anywhere
\end{lstlisting}
The same recipe transfers to other systems for which an approximate
$f_\text{phys}$ can be written --- for example Lorenz-63,
FitzHugh--Nagumo, or chemical-kinetics models --- with a conserved
quantity required only for the Hamiltonian-drift axis.

\section{Conclusion}
\label{sec:conclusion}

We introduced \mpinode, a Neural ODE framework that combines a soft
physics-informed residual with a Multiple-Initial-Condition
multiple-shooting curriculum. The case for combining the two is
structural rather than additive: each ingredient's blind spot is
exactly the regime the other constrains. On Lotka--Volterra,
\mpinode\ achieves the lowest OOS and long-horizon error among
purely data-driven methods and the lowest Hamiltonian drift, making
it the strongest method across all three axes simultaneously. In the typical
practitioner regime --- partial mechanistic knowledge, limited data
coverage, and a desire for long-horizon extrapolation --- this
makes \mpinode\ the most practical surrogate among the methods we
evaluated.

\paragraph{Generality beyond Lotka--Volterra.}
The construction is system-agnostic: any dynamical system for which
an approximate $f_\text{phys}$ can be written fits into the same
training and evaluation pipeline, with a conserved quantity required
only for the Hamiltonian-drift axis. Many systems of practical
interest --- dissipative or chaotic ones, such as a damped
oscillator --- admit no conserved Hamiltonian at all. For
these, the method itself and its other two axes (out-of-sample
accuracy and long-horizon stability) apply unchanged; the
conservation axis is simply replaced by whatever invariant or
stability diagnostic the system does admit --- a known first
integral, an energy or mass budget, or a Lyapunov-type bound --- or
omitted entirely when none exists. The conservation axis is
therefore a system-dependent diagnostic, not a precondition for
applying \mpinode. The framework applies to systems satisfying four
conditions: (i) \emph{autonomous} or weakly time-driven, so that the
Neural-ODE adjoint is well-posed; (ii) \emph{continuous in state}, so
that the adaptive integrator does not encounter switching events;
(iii) \emph{smooth enough for adaptive Runge--Kutta integration}, so
that stiff regions do not exhaust the time budget; and (iv) \emph{at
least partial mechanistic knowledge} expressible as $f_\text{phys}$.
When no partial physics is available the method still applies with
$\lambda_\text{phys} = 0$ (MIC-only), though the strongest guarantees
require both ingredients. Systems outside this scope might include
those with discrete switching and strongly forced or controlled
dynamics.

\paragraph{Practical guidance and broader impact.}
The framework's soft-constraint design lets the learner override
assumed physics when data contradicts it; this is a feature for
scientific exploration but a risk in safety-critical settings, where
mis-specified $f_\text{phys}$ could silently bias forecasts.
Practitioners deploying \mpinode\ in such settings should report
all three evaluation axes proposed here --- out-of-sample accuracy,
long-horizon stability, and conservation-law fidelity --- rather
than a single trajectory-error scalar, so that silent geometric
failures of the learned vector field remain visible to auditors.
Direct applications include ecological forecasting, chemical
kinetics modeling, and epidemic spread modeling; in each case the
three-axis report makes the typical failure modes of a learned
surrogate auditable in a way that a single MSE cannot.

\begin{acknowledgments}
The authors acknowledge the Department of Chemical Engineering,
Imperial College London, for computational resources and support.
\end{acknowledgments}

\section*{AUTHOR DECLARATIONS}

\subsection*{Conflict of Interest}

The authors have no conflicts of interest to disclose.

\subsection*{Author Contributions}

\textbf{Lake Yang}: Conceptualization (equal); Methodology (lead);
Software (lead); Investigation (lead); Writing -- original draft
(lead). \textbf{Antonio Malpica-Morales}: Methodology (supporting);
Writing -- review \& editing (supporting). \textbf{Frank Ioannis
Papadakis Wood}: Writing -- review \& editing (supporting).
\textbf{Serafim Kalliadasis}: Supervision (lead); Writing -- review
\& editing (supporting).

\section*{Data Availability Statement}

The data that support the findings of this study --- including
training scripts, hyperparameter configurations, and reproduction
artifacts for every figure and table --- are openly available in the
\textsc{MPINeuralODE} GitHub repository at
\url{https://github.com/LakeYang0818/MPINeuralODE},
Ref.~\onlinecite{MPINeuralODErepo}, under the MIT license.

\appendix

%% ───────────────────────────────────────────────────────────
%%  APPENDIX A — Architecture Search
%% ───────────────────────────────────────────────────────────
\section{Architecture Search Details}
\label{app:architecture}

This appendix collects per-configuration diagnostics from the
Phase-1 architecture grid search referenced in \cref{sec:phase1}.
We swept network depth (2--6 hidden layers) and width ($32$, $64$,
$128$, $256$ units per layer) at fixed remaining hyperparameters,
with the composite metric (log-space mean of in-sample, OOS, and
long-horizon MSE) on a fixed validation set.

The pattern that emerged is that depth and width contribute
non-trivially up to $\sim\!4$ hidden layers and $\sim\!128$ units,
beyond which validation error plateaus and additional capacity does
not pay for itself --- wider and deeper networks converge to the
same long-horizon error within the seed-to-seed band while costing
proportionally more solver evaluations per integration step. The
\textbf{$128 \times 4$ tanh} configuration is therefore the favored
configuration of this sweep: it is the smallest network at the
plateau and the one \mpinode\ inherits. We retain it across every
method in this paper for the reasons set out in \cref{sec:phase1}.

\begin{figure}[tbp]
\centering
\includegraphics[width=\columnwidth]{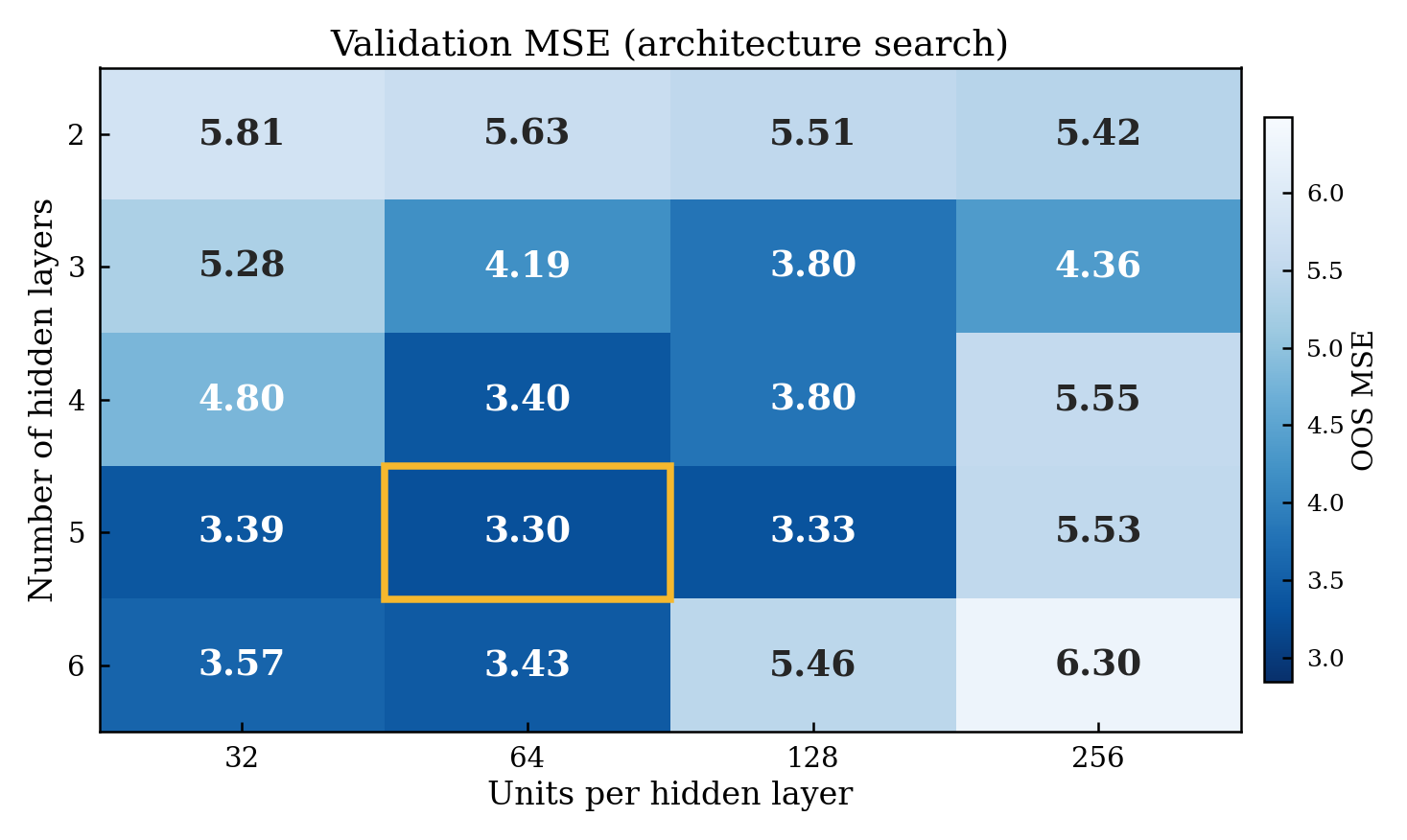}
\caption{\textbf{Architecture search heatmap.} Composite-metric
validation error across the depth--width grid. The $128 \times 4$
cell is the smallest network at the plateau and is adopted as the
default for every method in this paper.}
\label{fig:arch_heatmap}
\end{figure}

%% ───────────────────────────────────────────────────────────
%%  APPENDIX B — Temporal Sampling
%% ───────────────────────────────────────────────────────────
\section{Temporal Sampling Strategy Details}
\label{app:temporal}

\begin{figure*}[tbp]
\centering
\setlength{\tabcolsep}{3pt}
\renewcommand{\arraystretch}{1.05}
\begin{tabular}{|l|c|c|c|}
\hline
 & \footnotesize\textbf{OOS TS} & \footnotesize\textbf{OOS phase} & \footnotesize\textbf{Training} \\[1pt]
\hline
\footnotesize\textbf{Full window} &
  \includegraphics[width=0.31\textwidth]{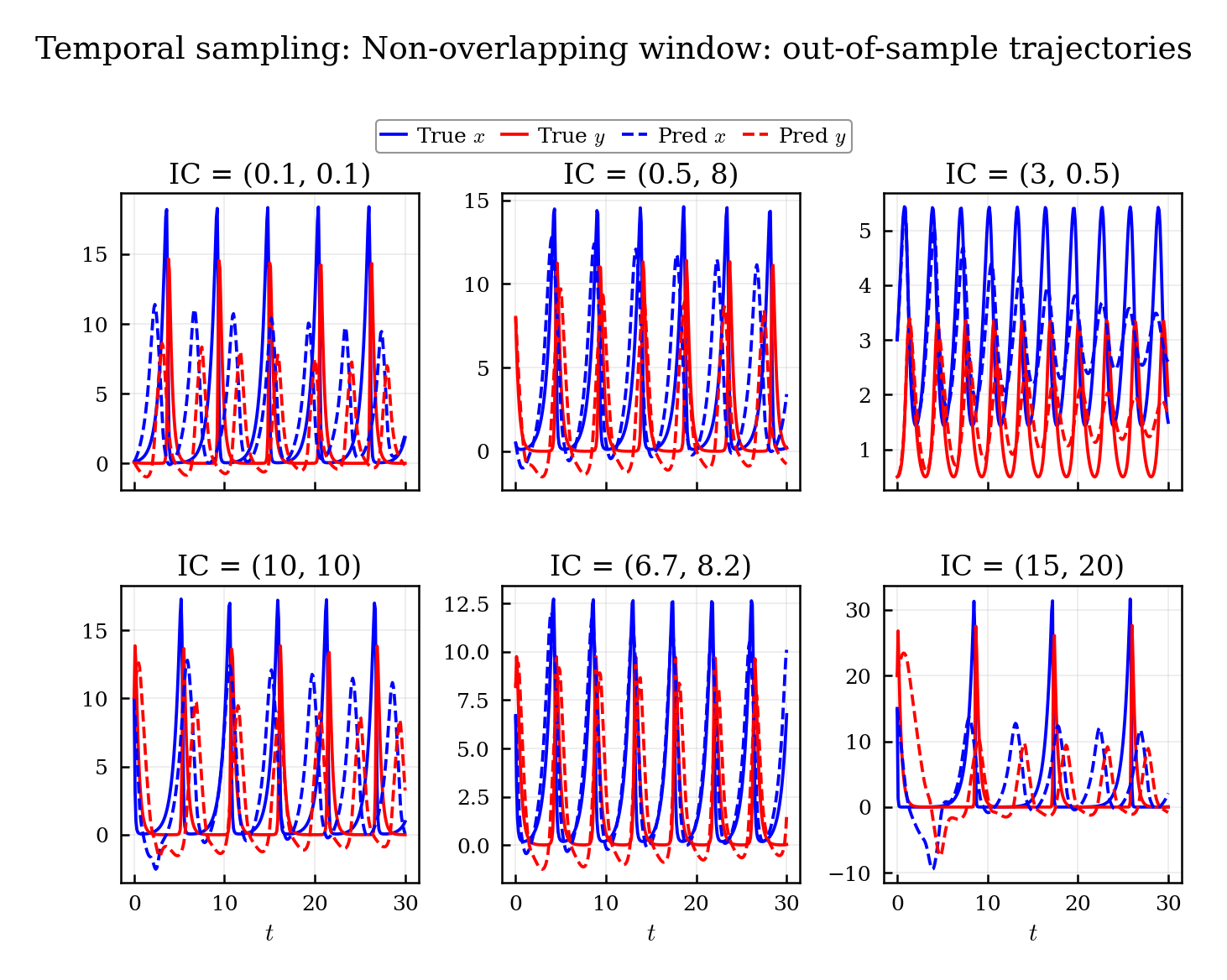} &
  \includegraphics[width=0.22\textwidth]{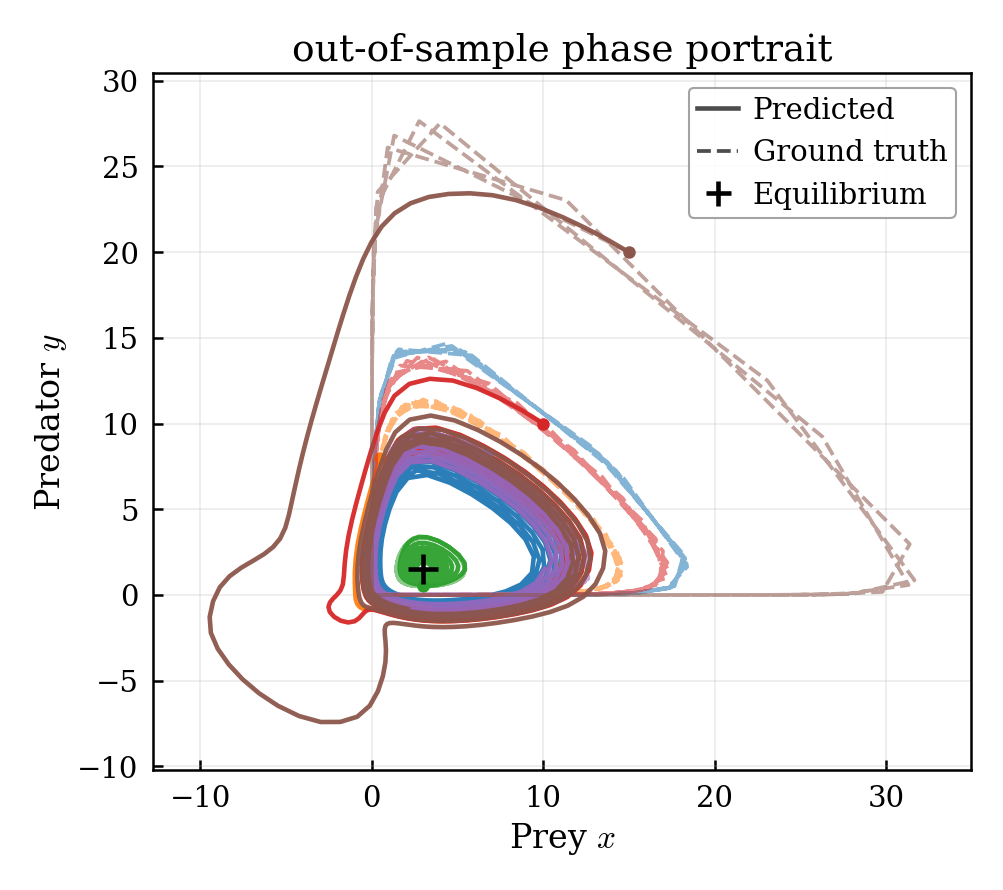} &
  \includegraphics[width=0.31\textwidth]{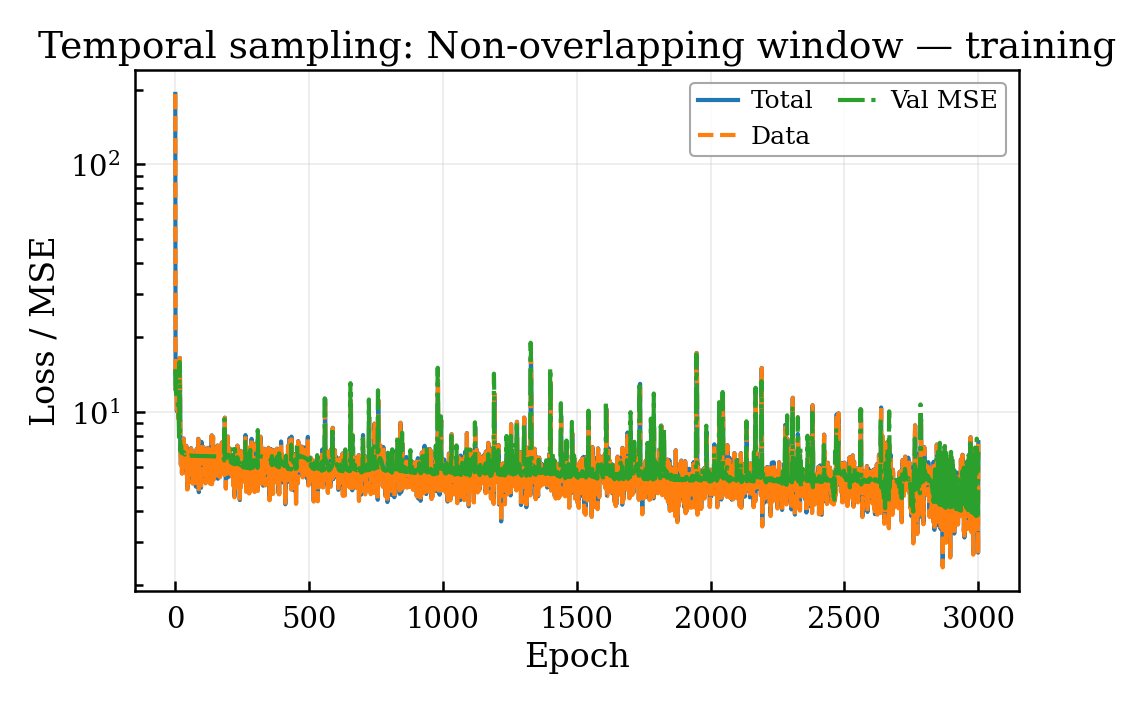} \\[1pt]
\hline
\footnotesize\textbf{Expanding} &
  \includegraphics[width=0.31\textwidth]{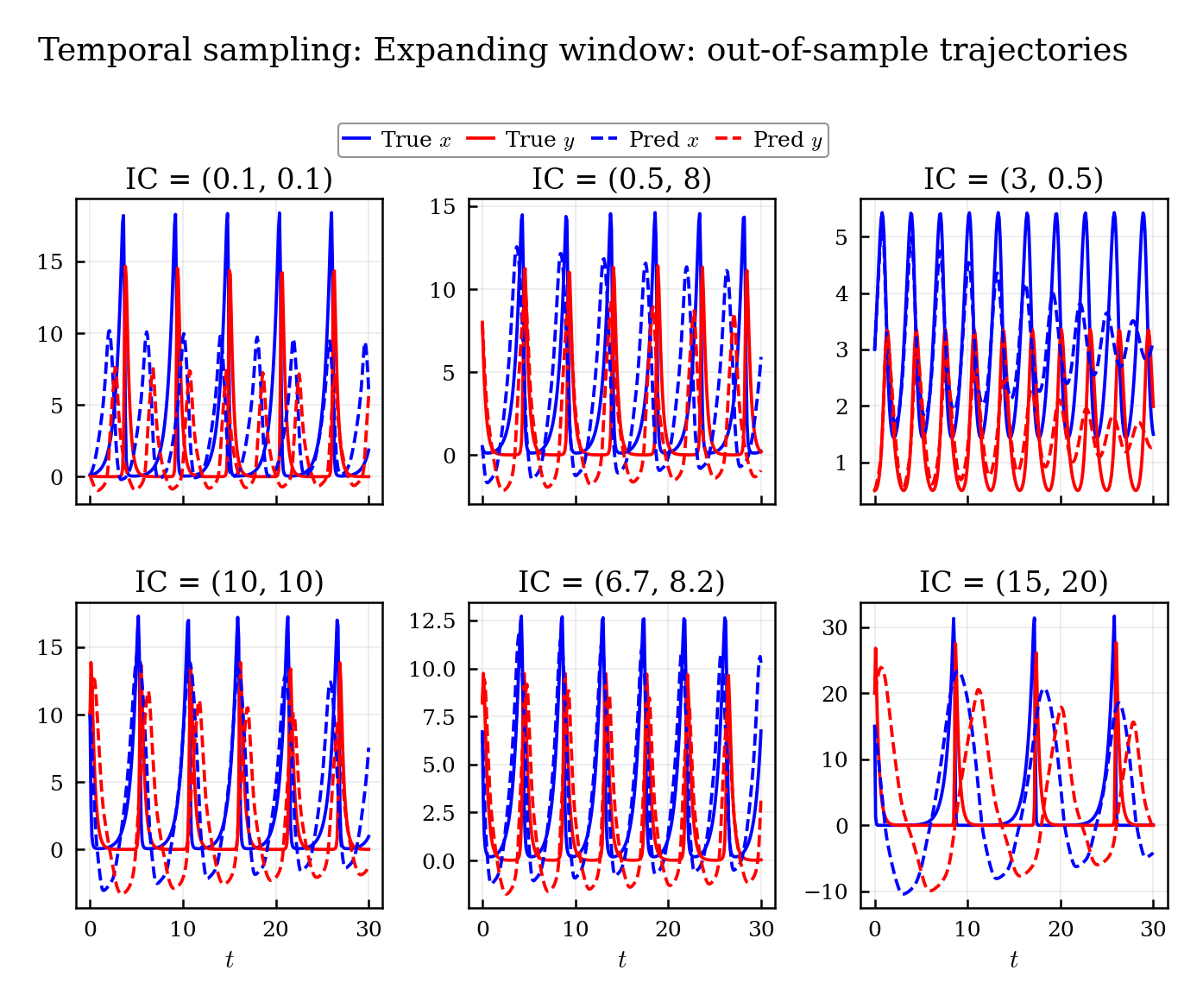} &
  \includegraphics[width=0.22\textwidth]{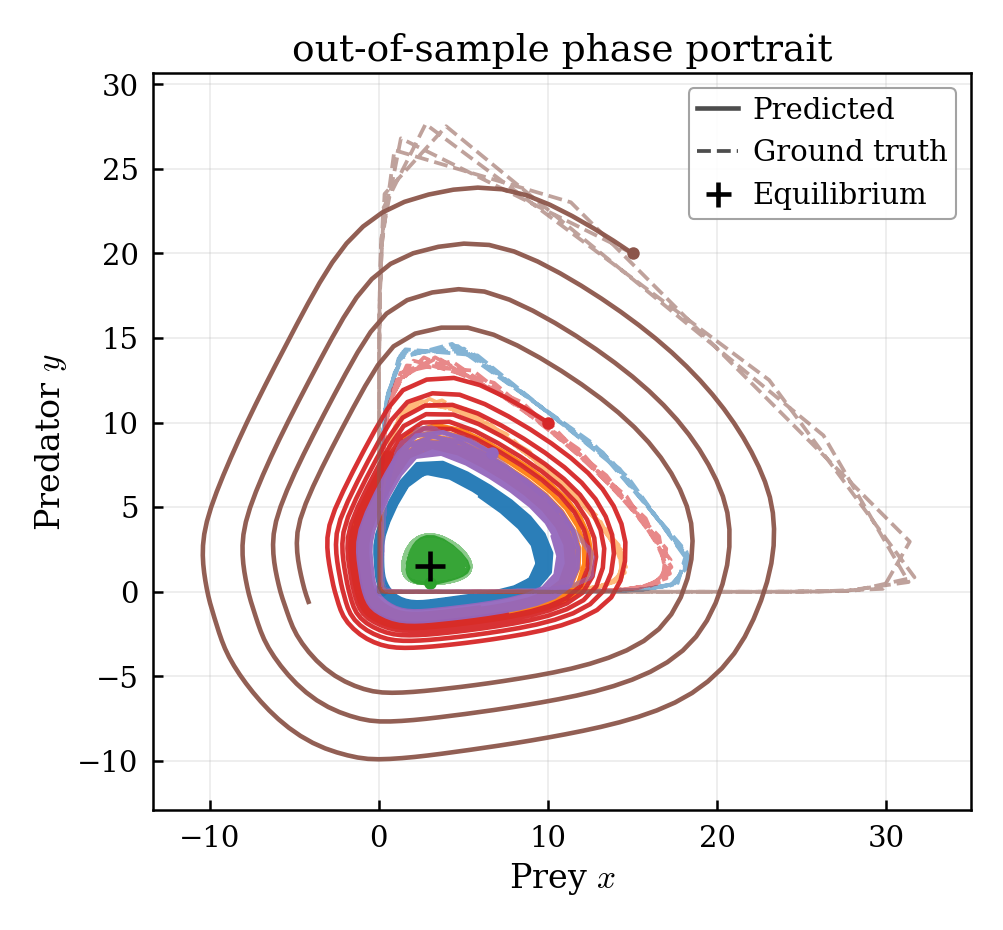} &
  \includegraphics[width=0.31\textwidth]{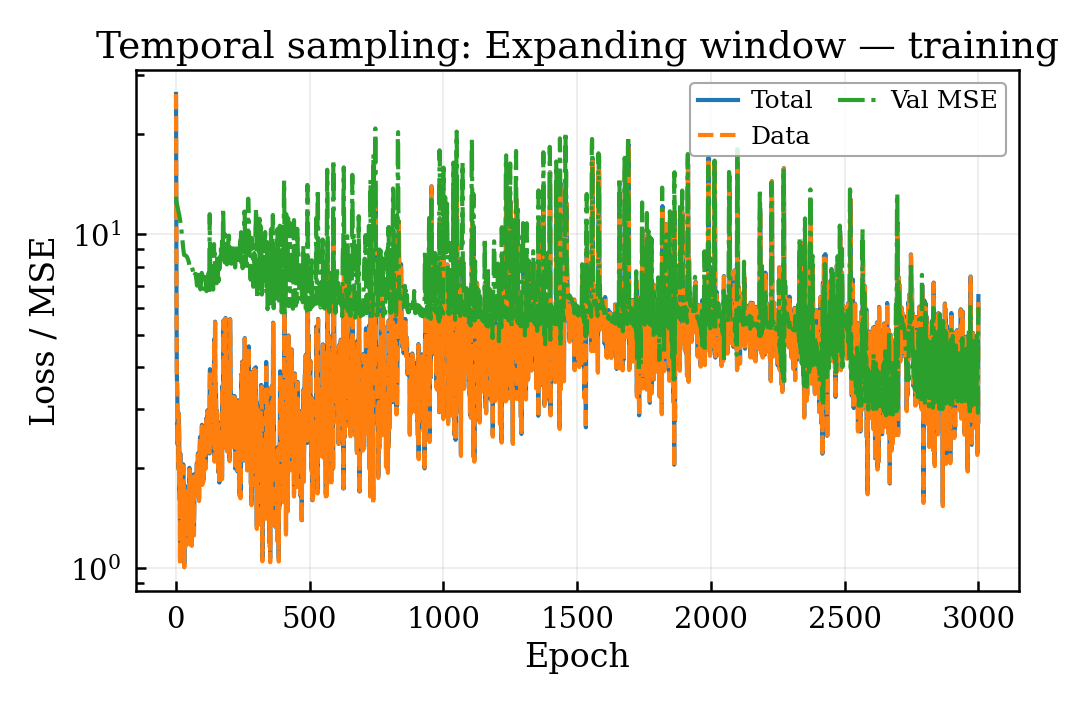} \\[1pt]
\hline
\footnotesize\textbf{Sliding} &
  \includegraphics[width=0.31\textwidth]{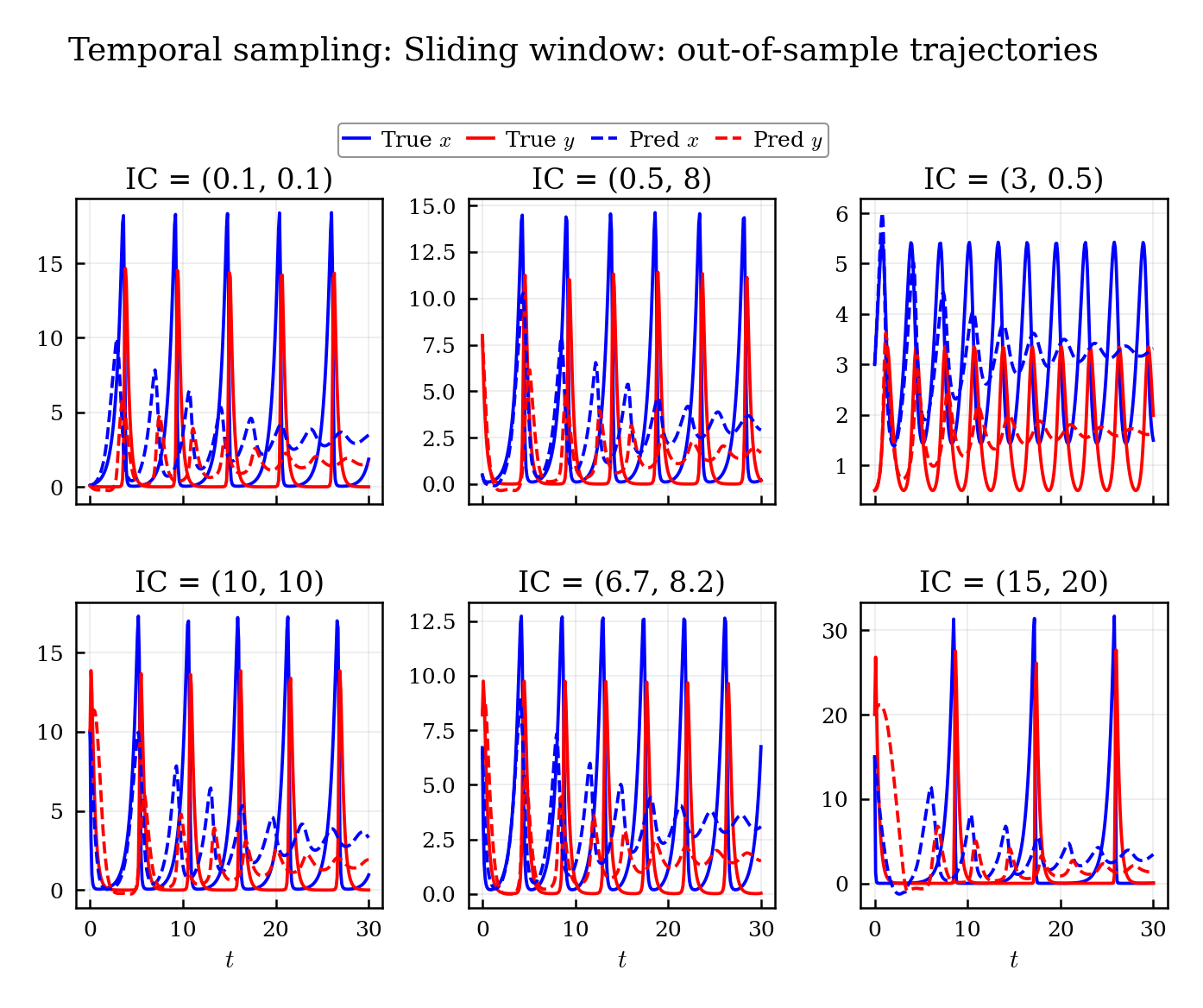} &
  \includegraphics[width=0.22\textwidth]{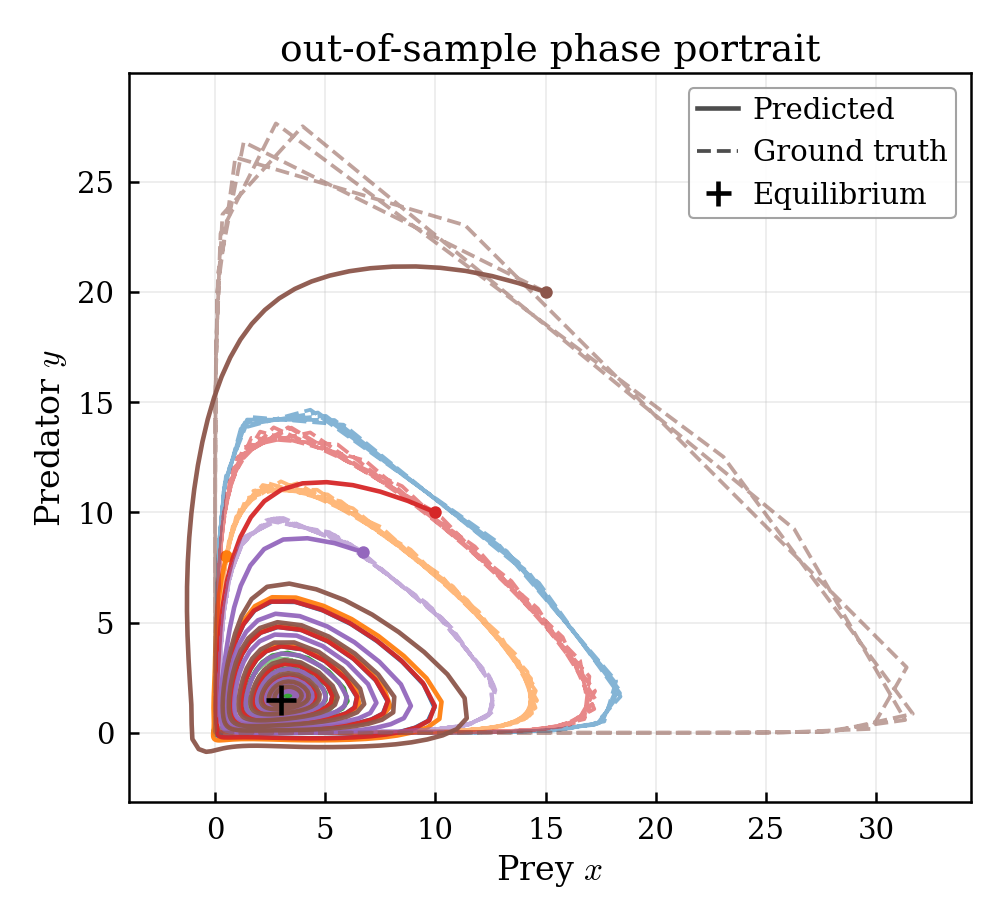} &
  \includegraphics[width=0.31\textwidth]{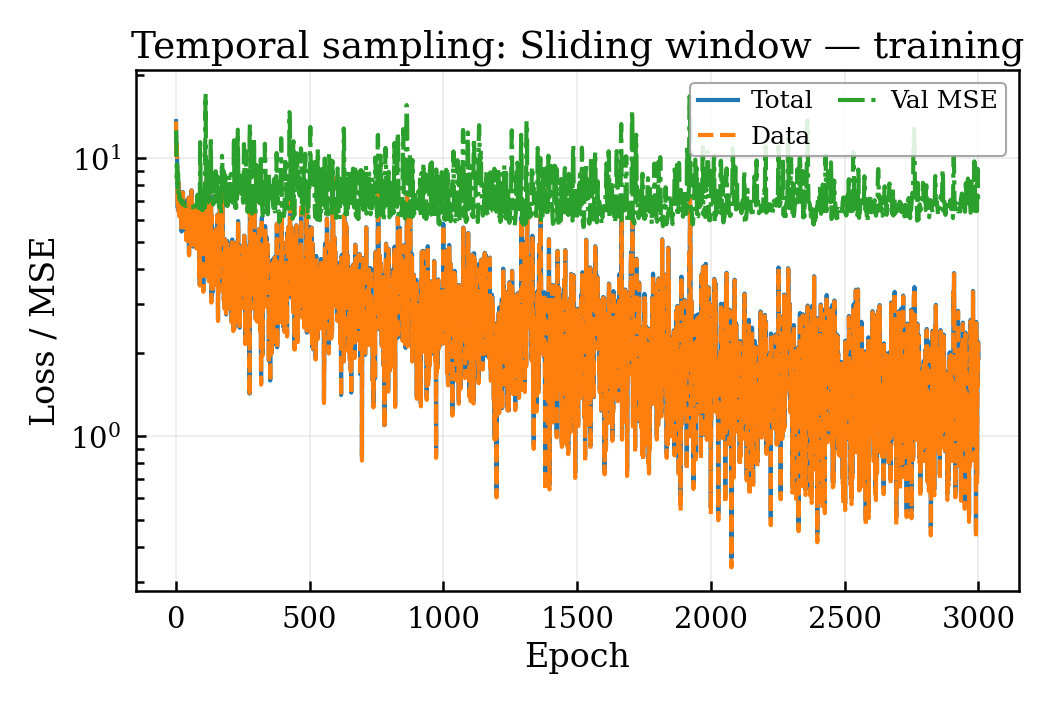} \\
\hline
\end{tabular}
\caption{\textbf{Temporal sampling strategy comparison.}
Per-strategy diagnostics for the three temporal sampling schemes
evaluated in \cref{sec:phase1}. Each row: out-of-sample time series,
out-of-sample phase portrait, and training convergence. The
full-window scheme (row 1) is the favored configuration adopted as
the default; the curriculum variants (rows 2--3) leave the
long-horizon regime under-trained.}
\label{fig:app_time}
\end{figure*}

This appendix provides per-strategy diagnostics for the temporal
sampling comparison in \cref{sec:phase1}. We compared the
\emph{full-window} scheme, the \emph{expanding-window}
curriculum~\cite{Bengio2009Curriculum}, and \emph{sliding windows}
of fixed length $6.0$ at random offsets. The full-window scheme
achieves the lowest composite metric: paying the long-horizon cost
from epoch one drives the vector field toward consistency over the
full horizon from the outset, whereas the curriculum variants leave
the long-horizon regime under-trained when they finally expose it,
and the sliding-window scheme produces a noisier validation curve
because it commits to a fixed horizon length too early. The
qualitative comparison is summarized below; per-strategy
diagnostics (out-of-sample time series, phase portraits, training
curves) are shown in \cref{fig:app_time}.

%% ───────────────────────────────────────────────────────────
%%  APPENDIX C — Positivity Wrapper
%% ───────────────────────────────────────────────────────────
\section{Positivity Wrapper Details}
\label{app:positivity}

This appendix provides per-strategy diagnostics for the positivity
wrapper comparison in \cref{sec:phase1}. We evaluated four wrappers
on the raw output $r$ of the vector-field network: \emph{none}
(identity), \emph{tanh-bound} ($18\,\tanh(r/18)$), \emph{squared}
($r\,|r|$), and \emph{clamp} ($\textrm{clip}(r, -20, 20)$, which
truncates $r$ to the interval $[-20, 20]$).

\begin{figure*}[tbp]
\centering
\setlength{\tabcolsep}{3pt}
\renewcommand{\arraystretch}{1.05}
\begin{tabular}{|l|c|c|c|}
\hline
 & \footnotesize\textbf{OOS TS} & \footnotesize\textbf{OOS phase} & \footnotesize\textbf{Training} \\[1pt]
\hline
\footnotesize\textbf{Clamp} &
  \includegraphics[width=0.31\textwidth]{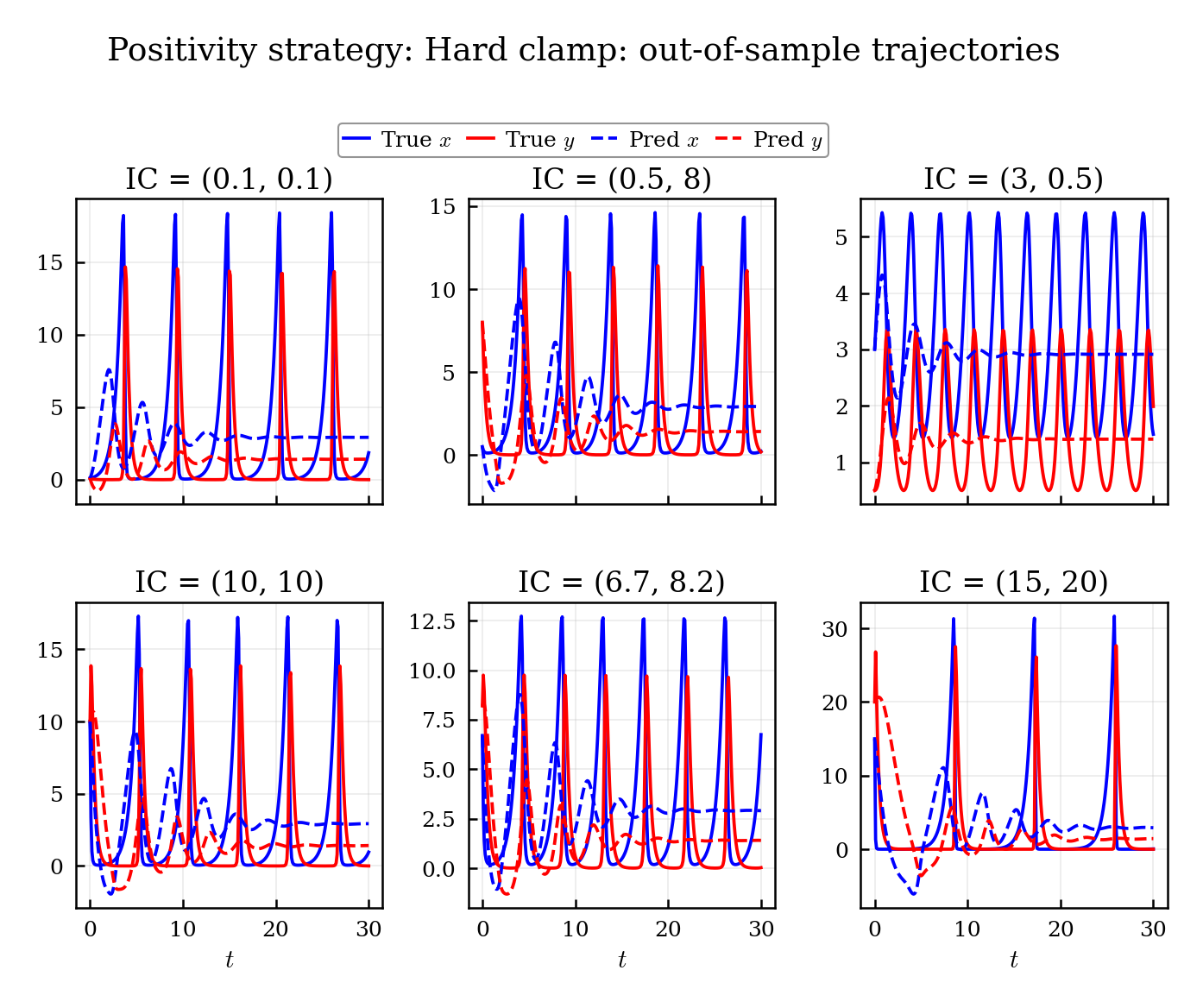} &
  \includegraphics[width=0.22\textwidth]{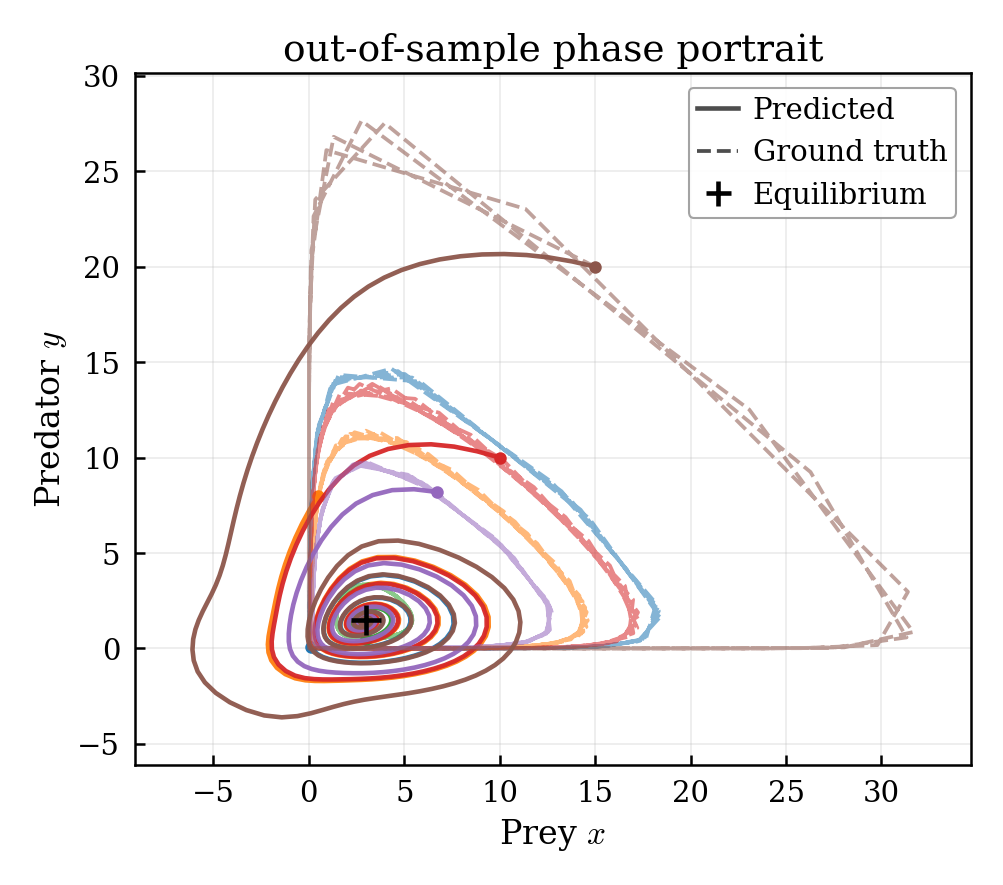} &
  \includegraphics[width=0.31\textwidth]{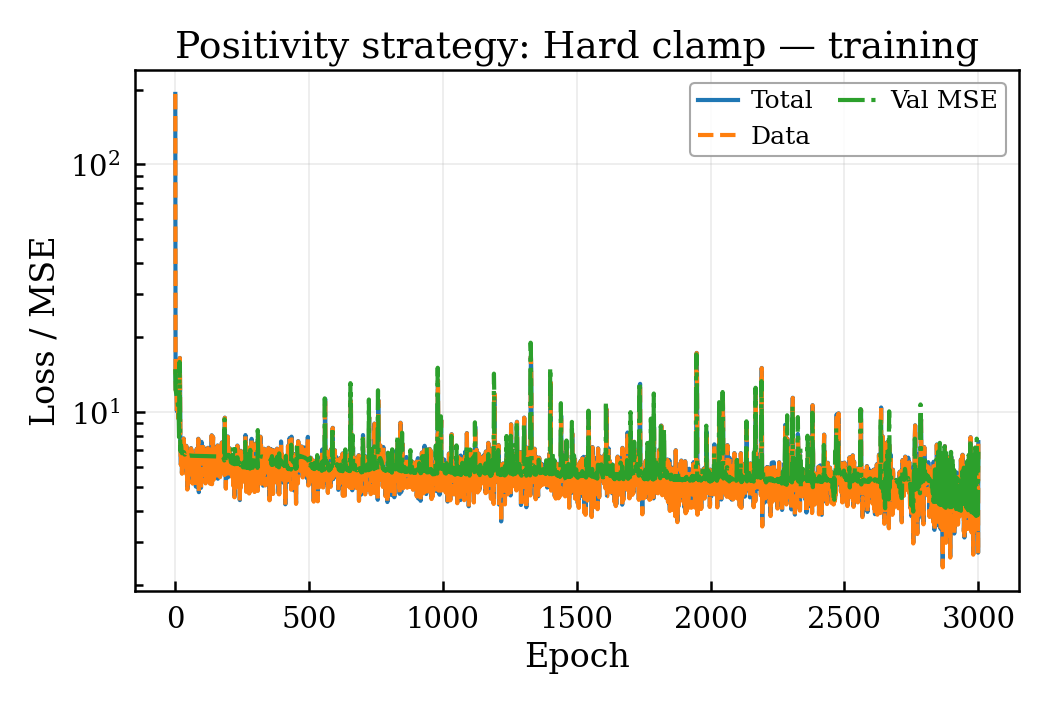} \\[1pt]
\hline
\footnotesize\textbf{Tanh-bound} &
  \includegraphics[width=0.31\textwidth]{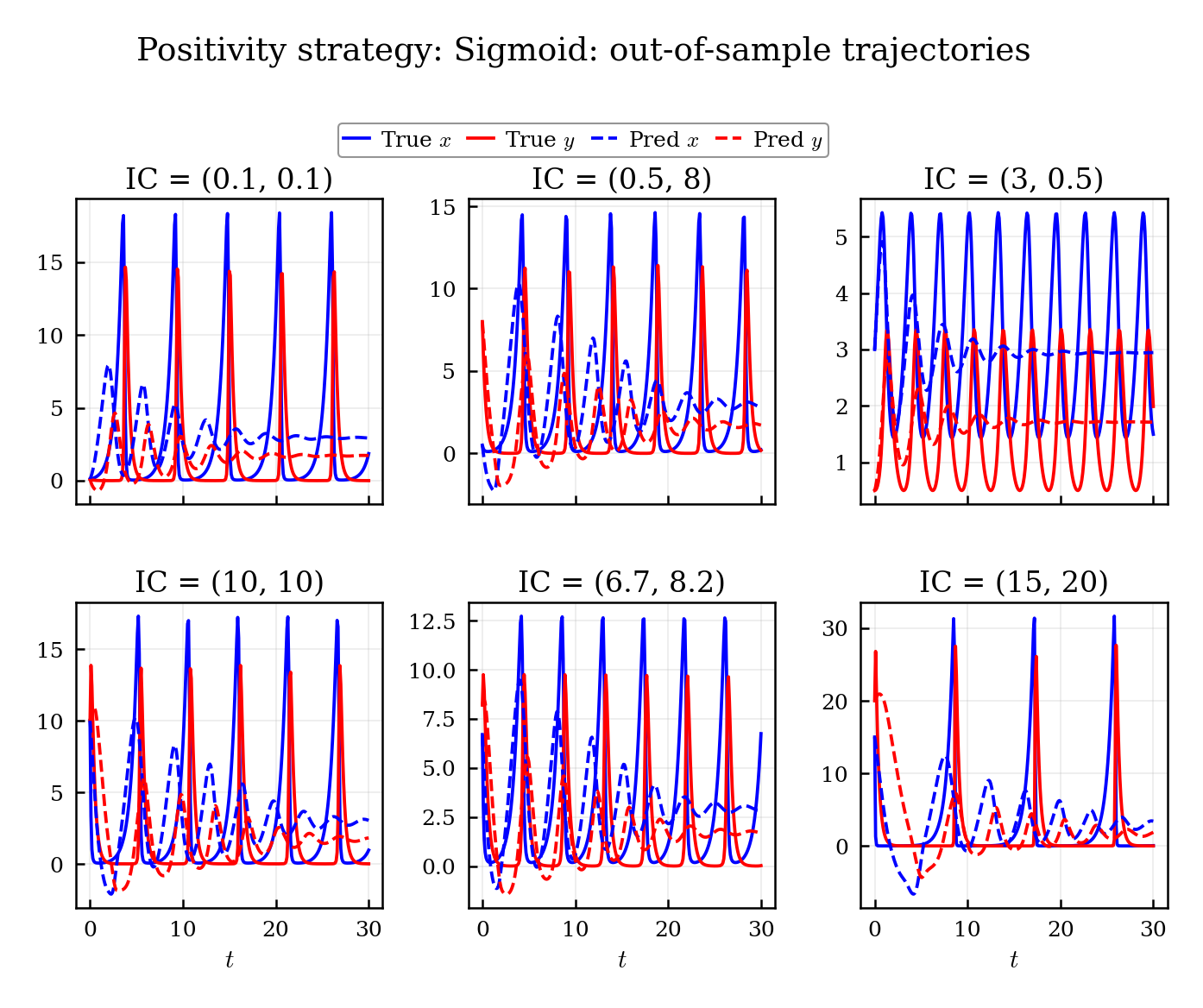} &
  \includegraphics[width=0.22\textwidth]{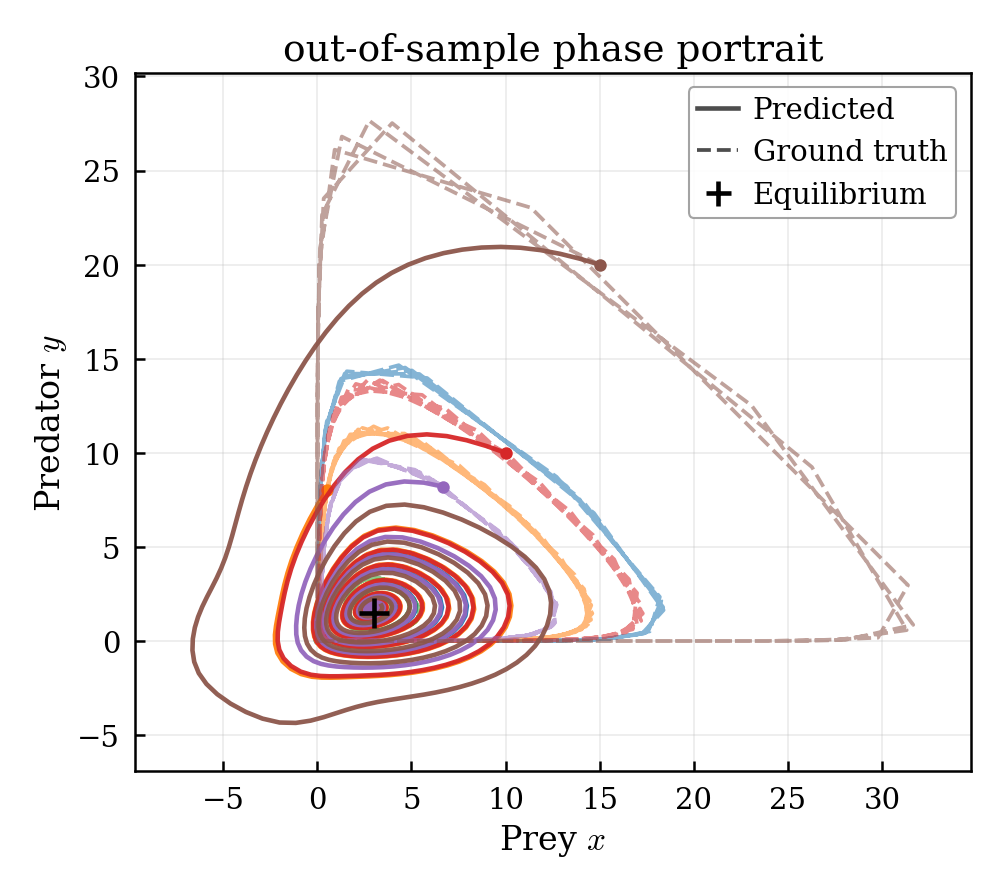} &
  \includegraphics[width=0.31\textwidth]{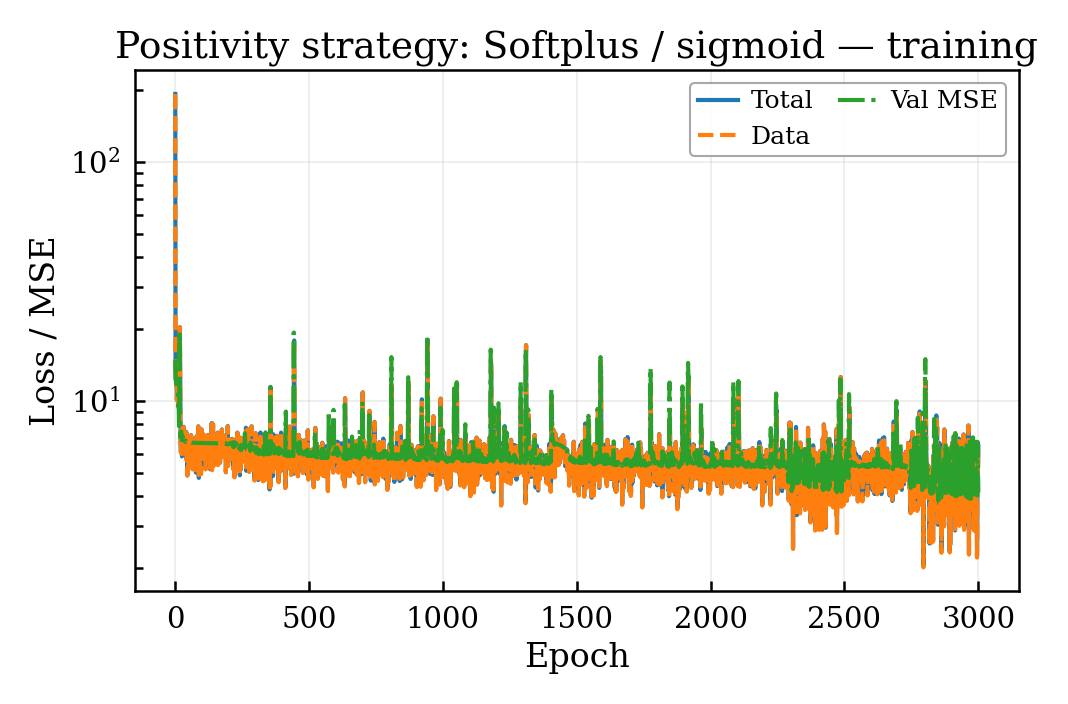} \\[1pt]
\hline
\footnotesize\textbf{Squared} &
  \includegraphics[width=0.31\textwidth]{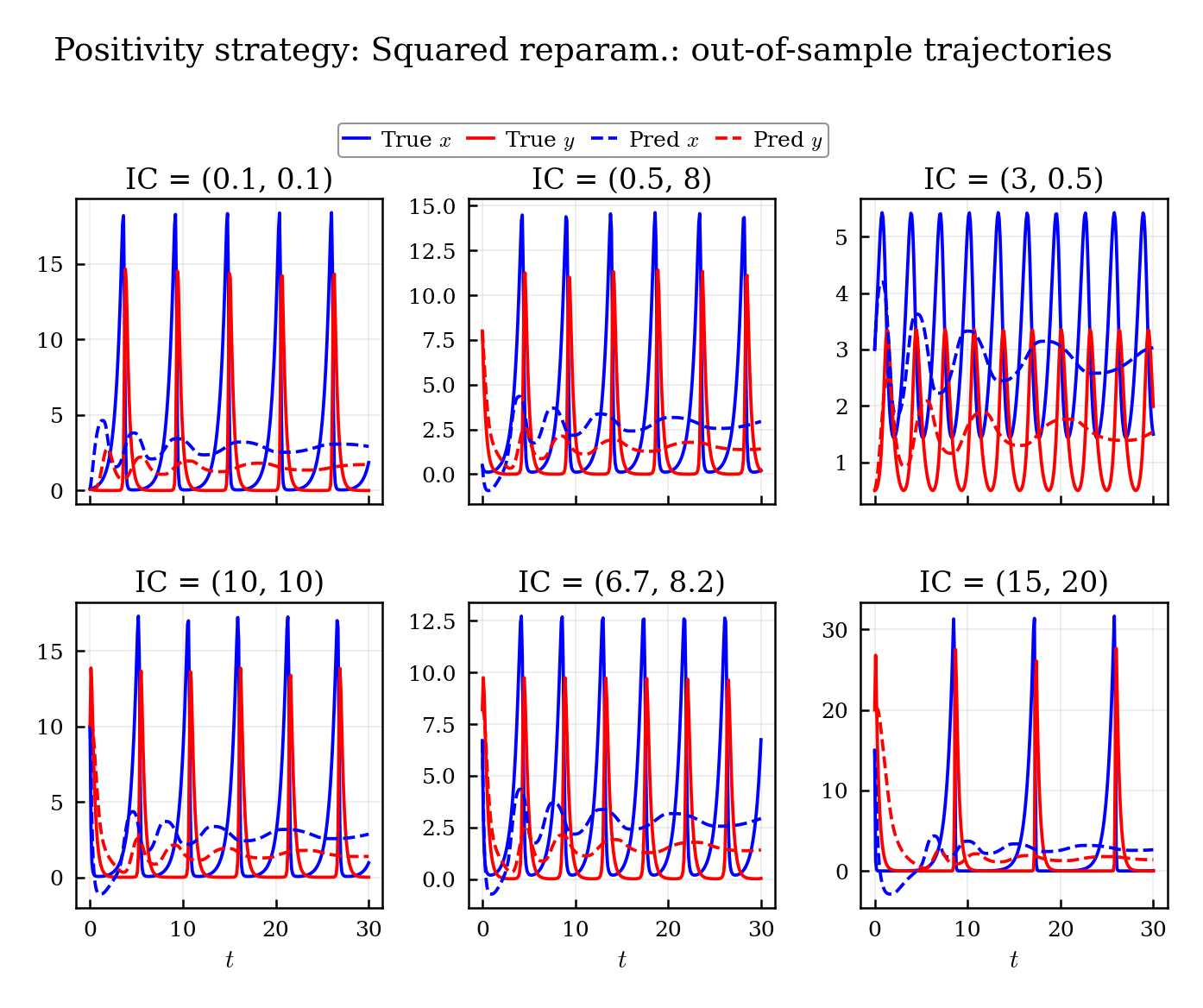} &
  \includegraphics[width=0.22\textwidth]{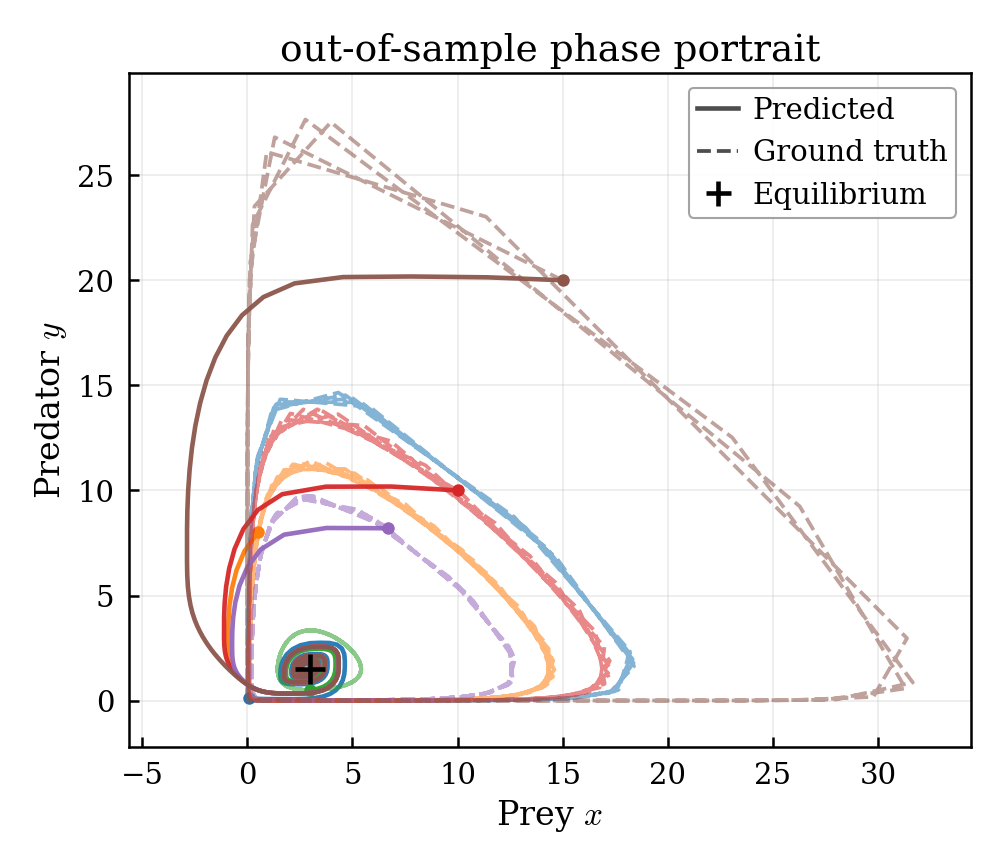} &
  \includegraphics[width=0.31\textwidth]{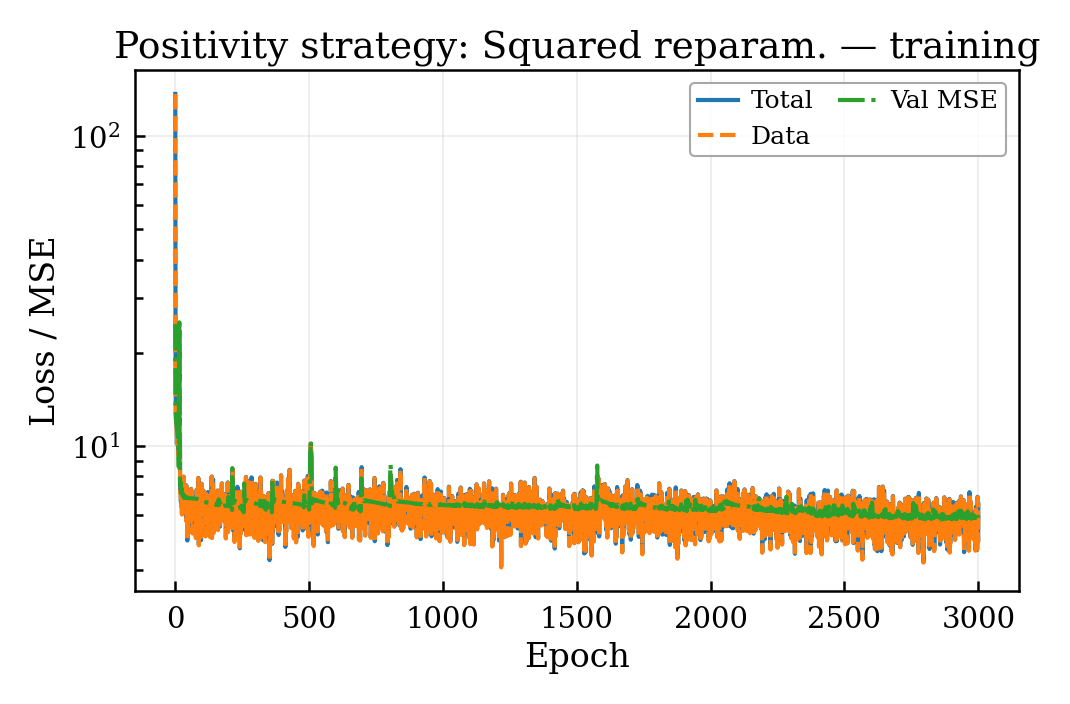} \\
\hline
\end{tabular}
\caption{\textbf{Positivity wrapper comparison.} Per-strategy
diagnostics for the wrappers evaluated in \cref{sec:phase1}. The
clamp wrapper (row 1) ties with the identity on the composite
metric and is adopted as the default deterministic safety net.
Tanh-bound (row 2) and squared (row 3) underperform either by
saturating useful signal or by distorting the derivative scale
near zero.}
\label{fig:app_neg}
\end{figure*}

The composite-metric scores for \emph{clamp} and \emph{none} agree
to ten significant digits: at width $128$ the raw network output
rarely approaches the $\pm 20$ clamp bound, so clamp acts as the
identity on virtually every forward pass and the two wrappers are
operationally identical on the typical training distribution. The
\emph{tanh-bound} wrapper saturates useful signal once the
vector-field magnitude approaches the bound, slightly slowing
convergence. The \emph{squared} wrapper distorts the derivative
scale near the origin --- $\partial(r |r|)/\partial r = 2|r|$
vanishes at $r = 0$ --- producing an early-training instability
visible in the training-curve panel of \cref{fig:app_neg}. We
adopt clamp because it gives a deterministic safety net against
the rare blow-up that the identity cannot prevent, at no measurable
cost on the typical-regime training distribution.

%% ───────────────────────────────────────────────────────────
%%  APPENDIX D — Per-Method Detailed Diagnostics
%%  (also resolves \cref{app:in_sample} from the empirical study)
%% ───────────────────────────────────────────────────────────
\section{Per-Method Detailed Diagnostics}
\label{app:method_details}\label{app:in_sample}

This appendix collects per-method in-sample fits, out-of-sample
diagnostics, and phase portraits for each of the five methods
compared in \cref{sec:three_axes_results}. The in-sample panels
illustrate the regularization tradeoff discussed in the body: the
unregularized baseline LV\_NN concentrates its $\sim\!67\text{k}$
parameters on a single typical-regime distribution and so attains
the tightest in-sample fit, while the regularized variants spread
the same capacity across a much broader sampler and accept a
slightly higher in-sample error in exchange for far better
out-of-sample and long-horizon behavior.

\begin{figure*}[tbp]
\centering
\setlength{\tabcolsep}{3pt}
\renewcommand{\arraystretch}{1.05}
\begin{tabular}{|l|c|c|c|}
\hline
 & \footnotesize\textbf{In-sample TS} & \footnotesize\textbf{OOS TS} & \footnotesize\textbf{OOS phase} \\[1pt]
\hline
\footnotesize\textbf{LV\_NN} &
  \includegraphics[width=0.31\textwidth]{plots/method_comparison/NN/time_series_in_sample} &
  \includegraphics[width=0.31\textwidth]{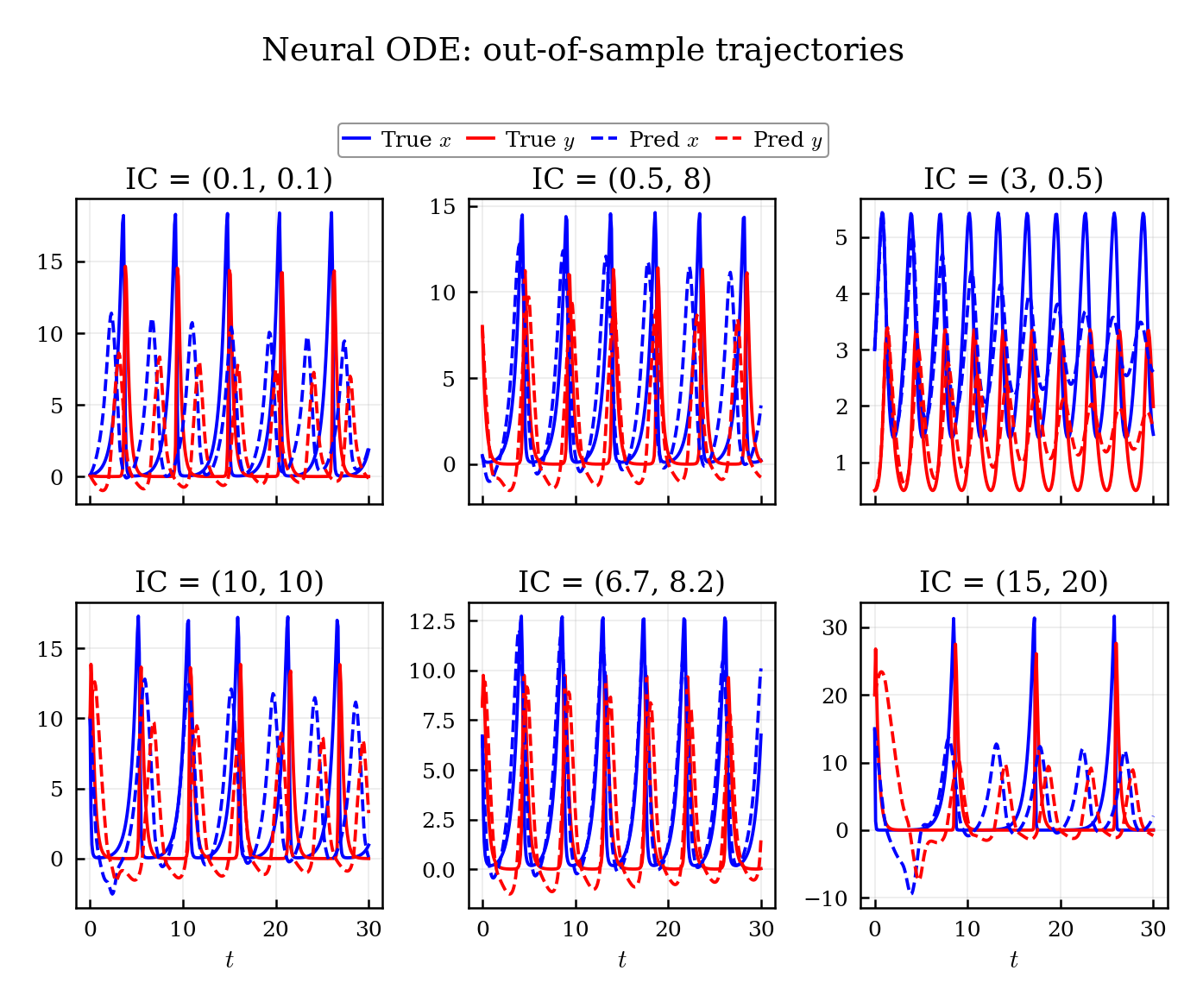} &
  \includegraphics[width=0.22\textwidth]{plots/method_comparison/NN/phase_oos} \\[1pt]
\hline
\footnotesize\textbf{LV\_PINN} &
  \includegraphics[width=0.31\textwidth]{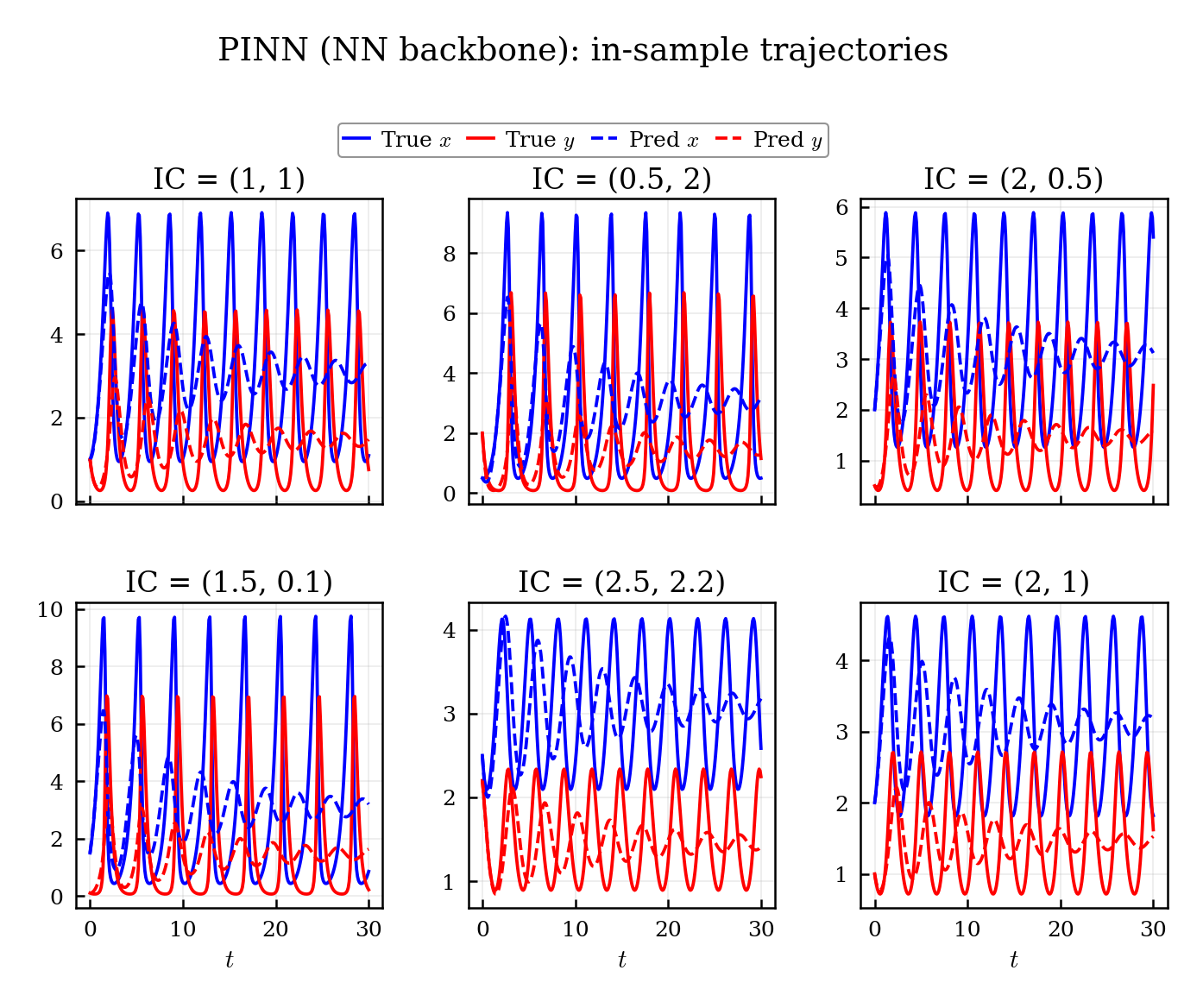} &
  \includegraphics[width=0.31\textwidth]{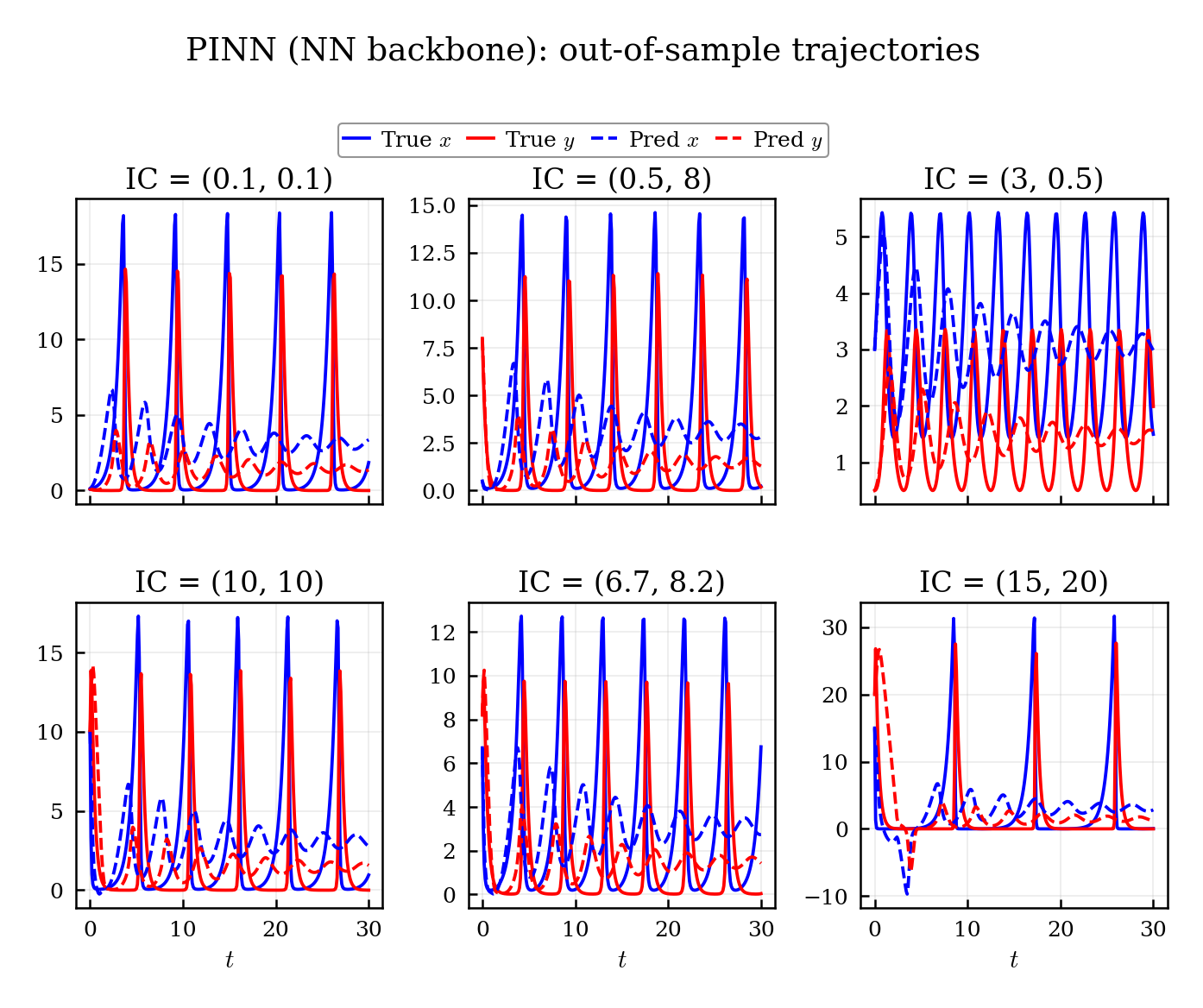} &
  \includegraphics[width=0.22\textwidth]{plots/method_comparison/PINN_NN/phase_oos} \\[1pt]
\hline
\footnotesize\textbf{LV\_MIC} &
  \includegraphics[width=0.31\textwidth]{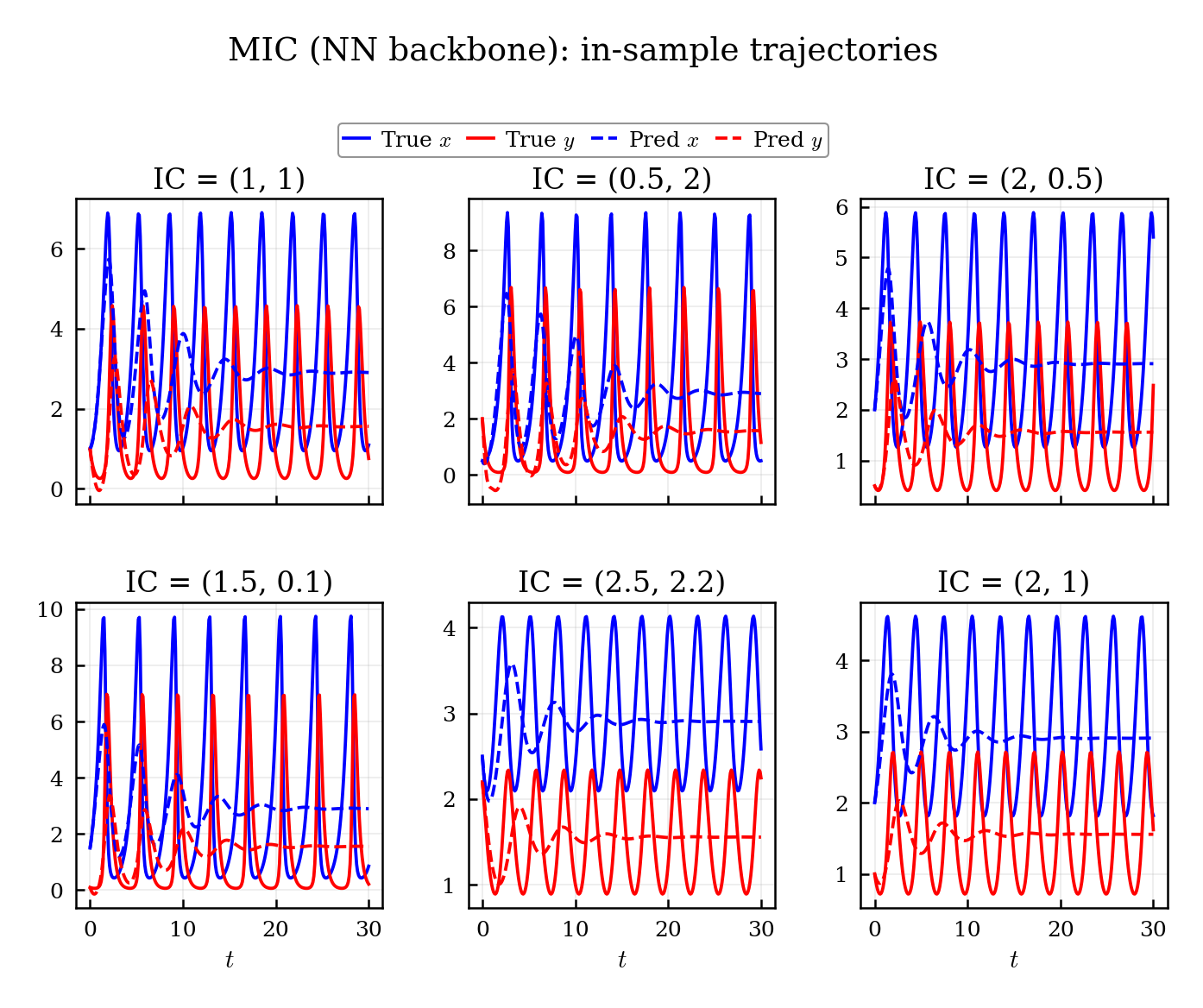} &
  \includegraphics[width=0.31\textwidth]{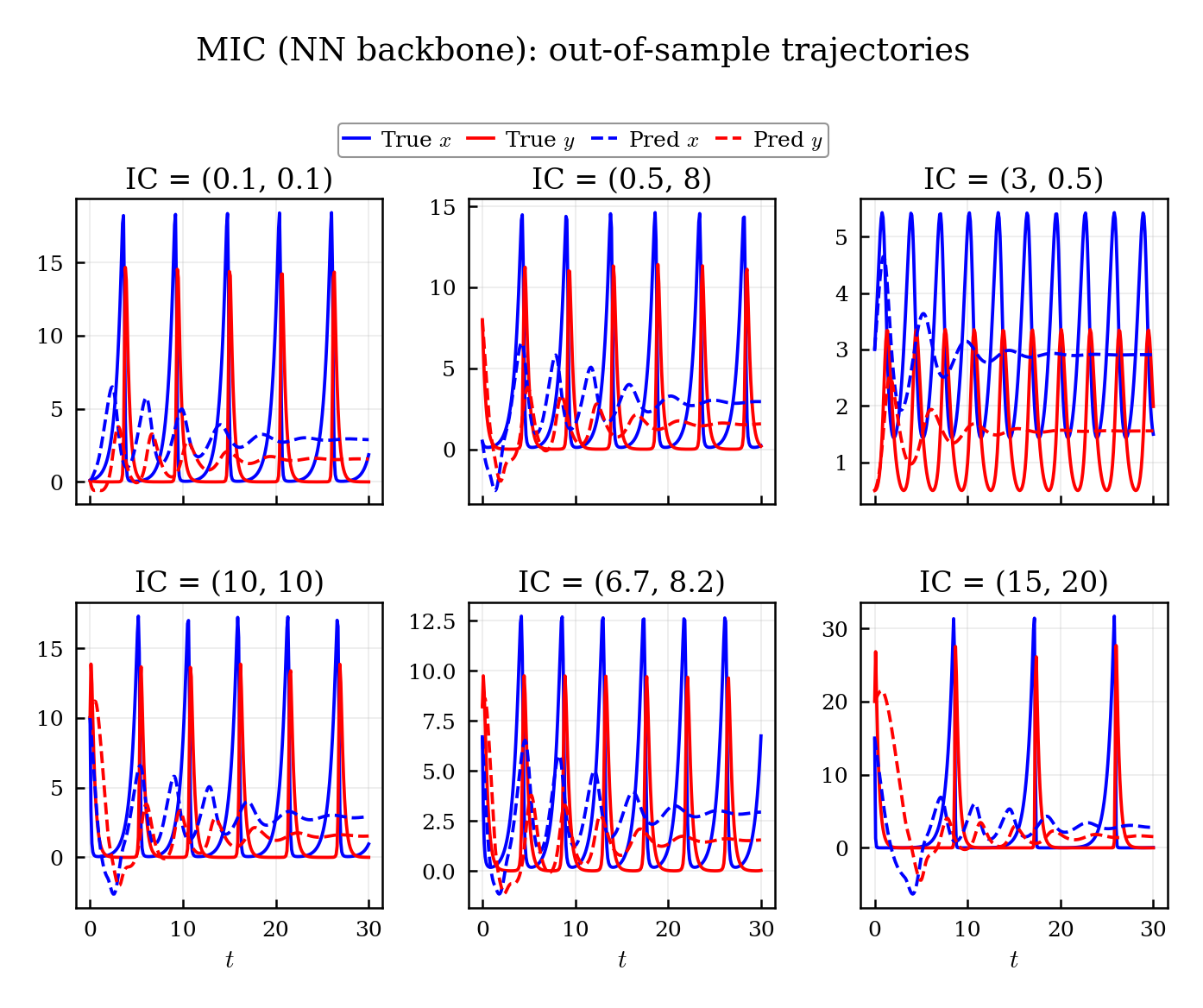} &
  \includegraphics[width=0.22\textwidth]{plots/method_comparison/MIC_NN/phase_oos} \\[1pt]
\hline
\footnotesize\textbf{\mpinode} &
  \includegraphics[width=0.31\textwidth]{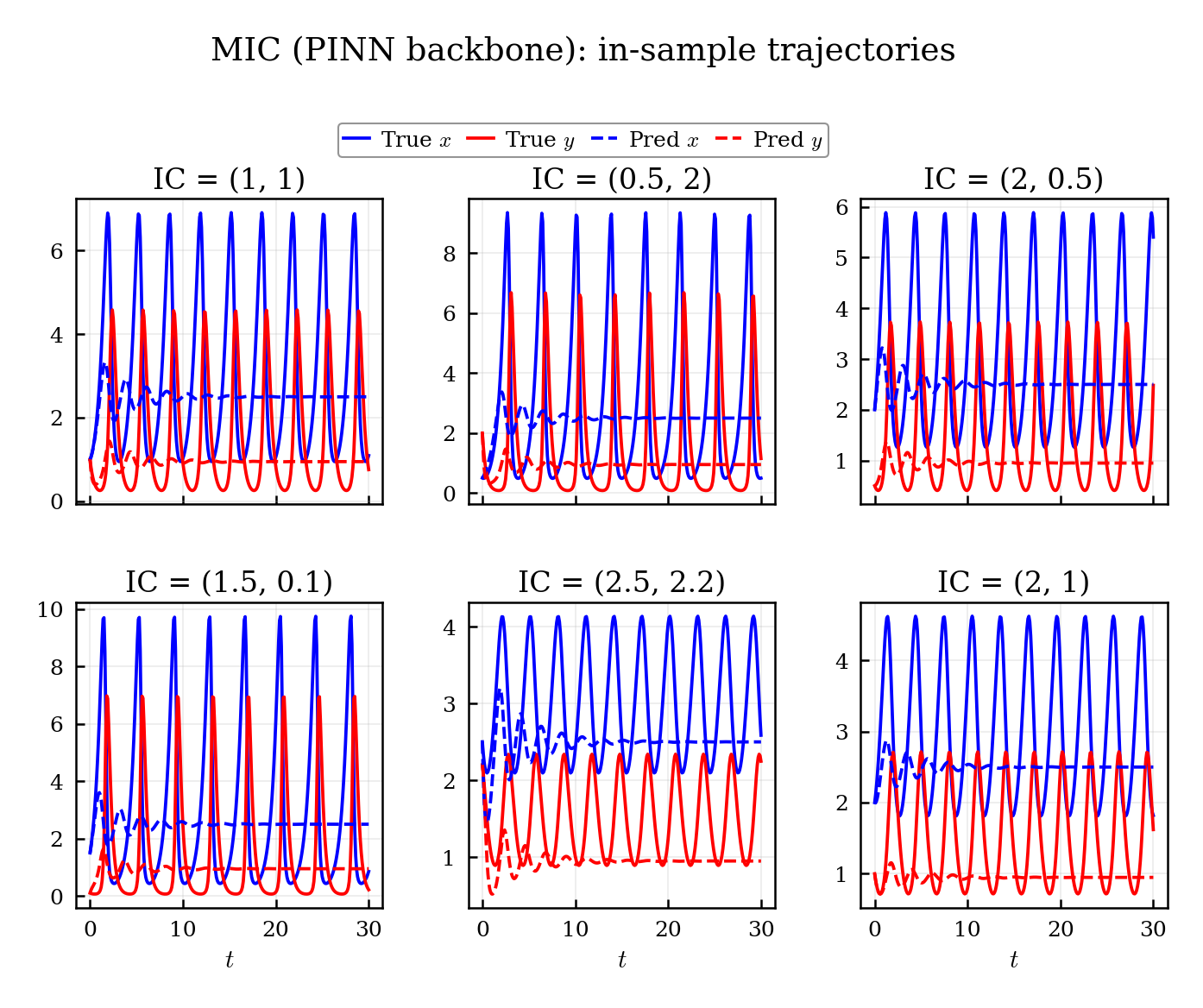} &
  \includegraphics[width=0.31\textwidth]{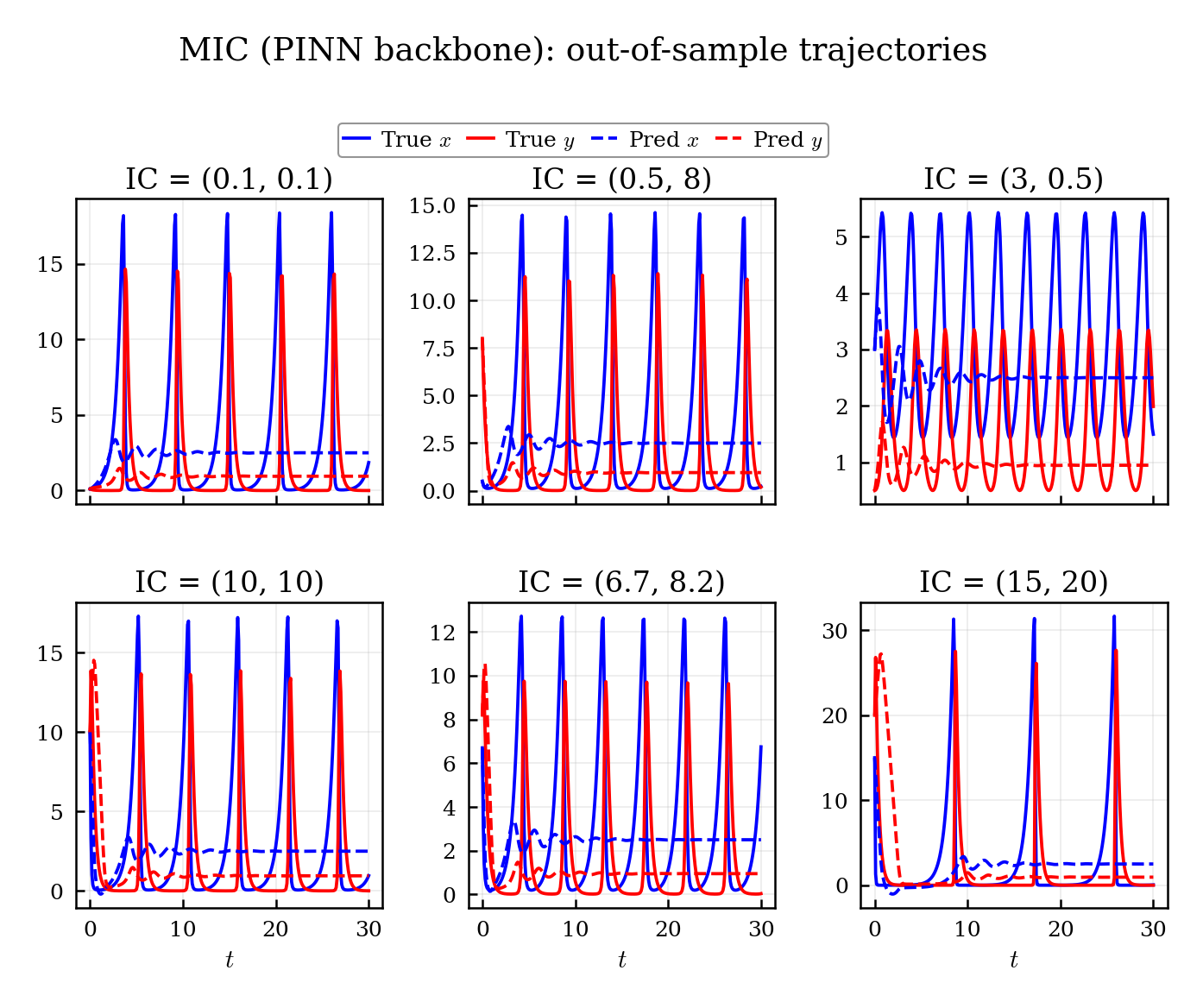} &
  \includegraphics[width=0.22\textwidth]{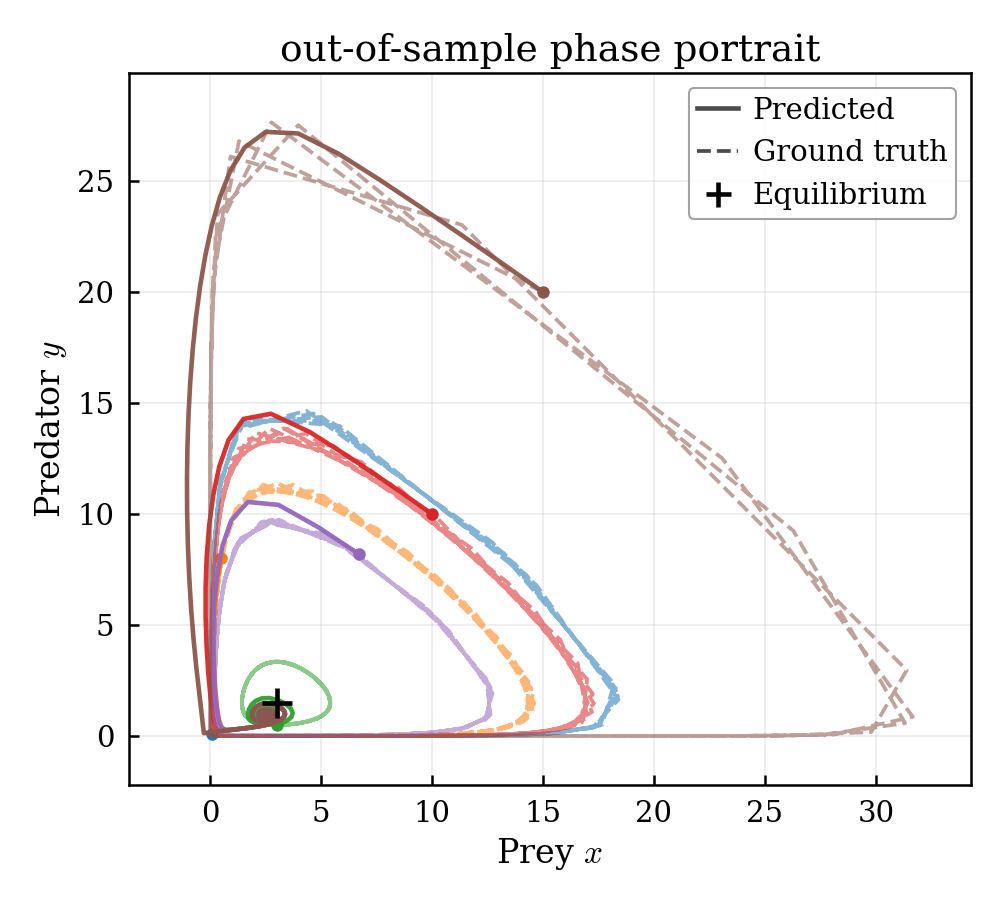} \\[1pt]
\hline
\footnotesize\textbf{LV\_Structured} &
  \includegraphics[width=0.31\textwidth]{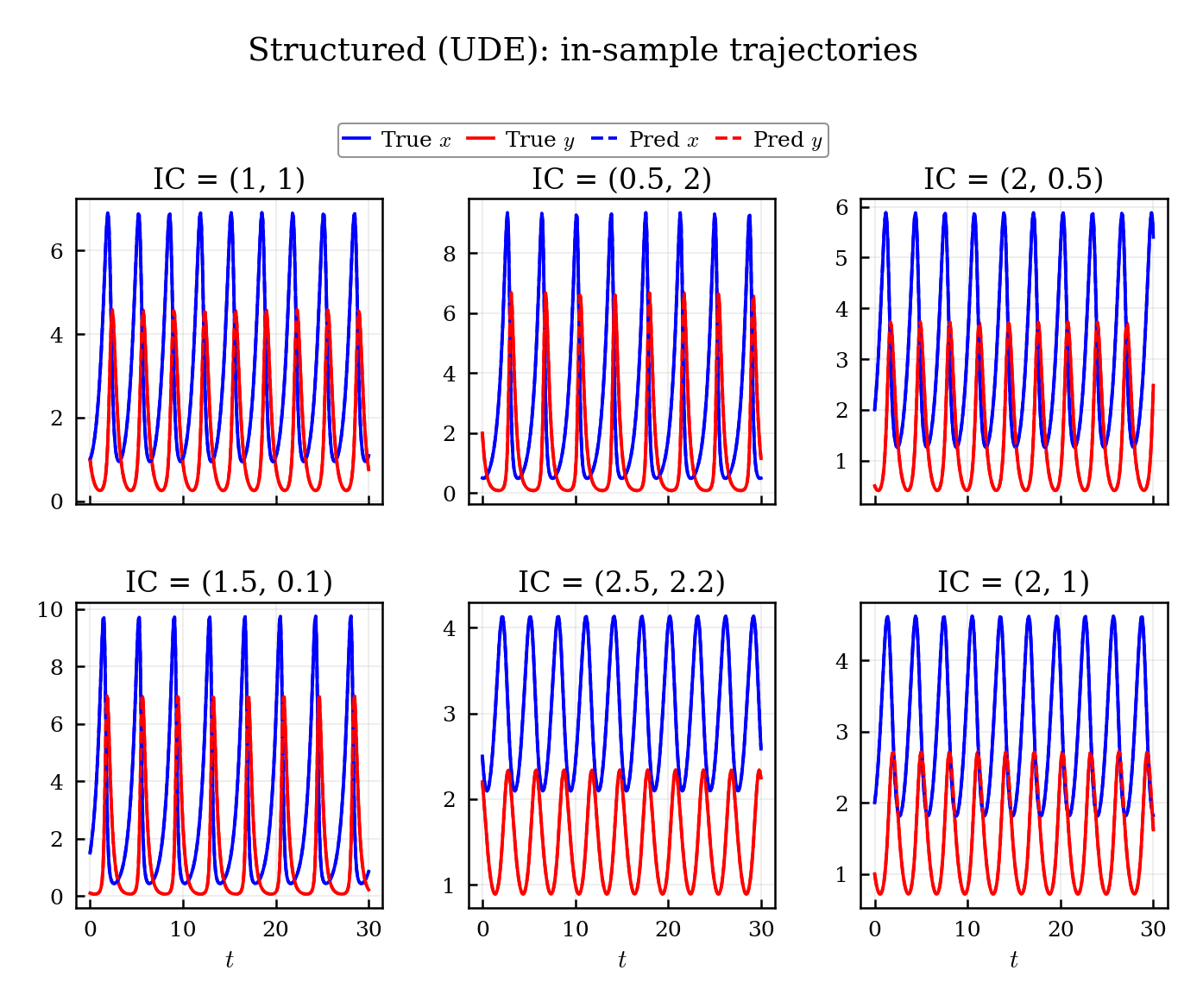} &
  \includegraphics[width=0.31\textwidth]{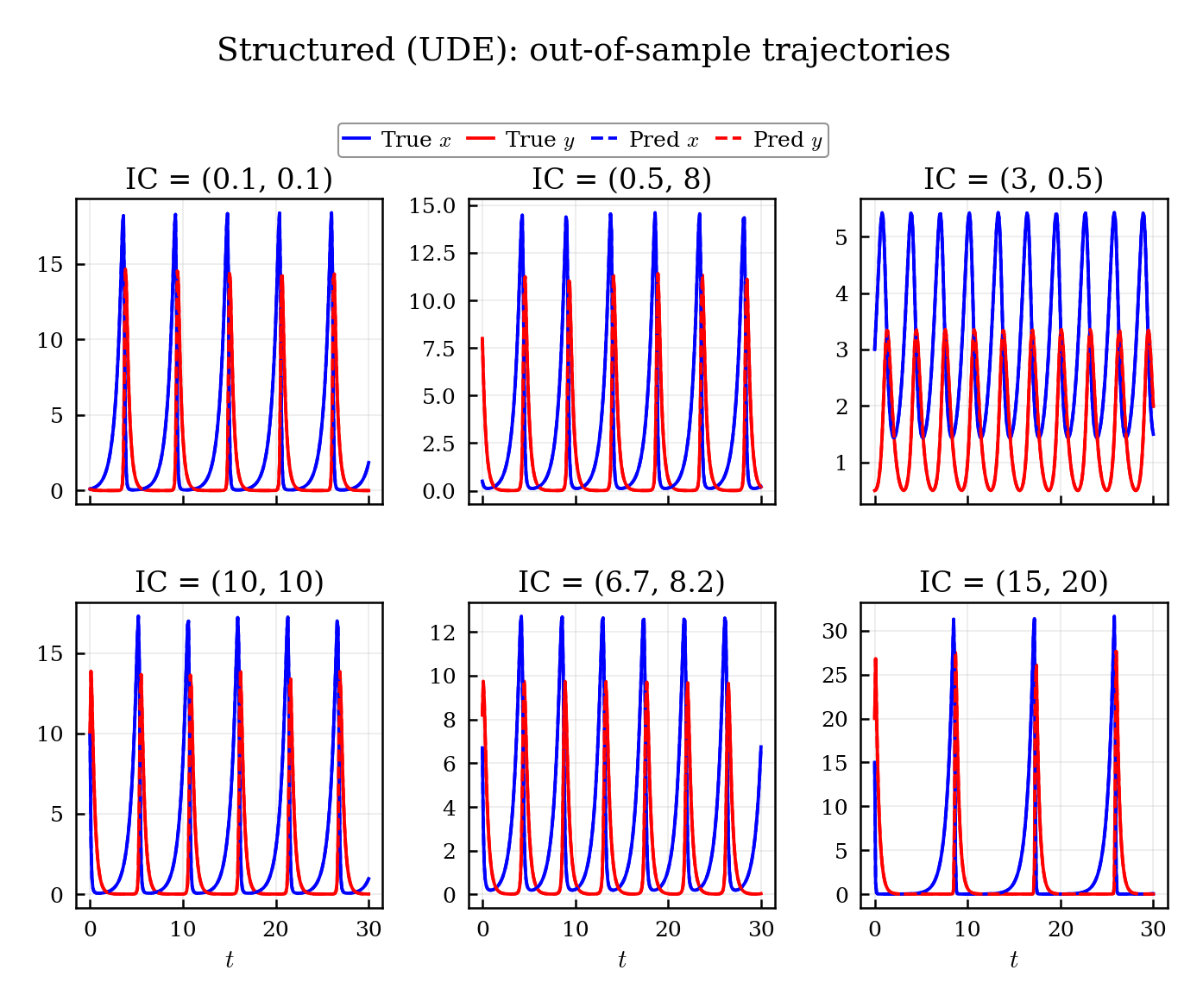} &
  \includegraphics[width=0.22\textwidth]{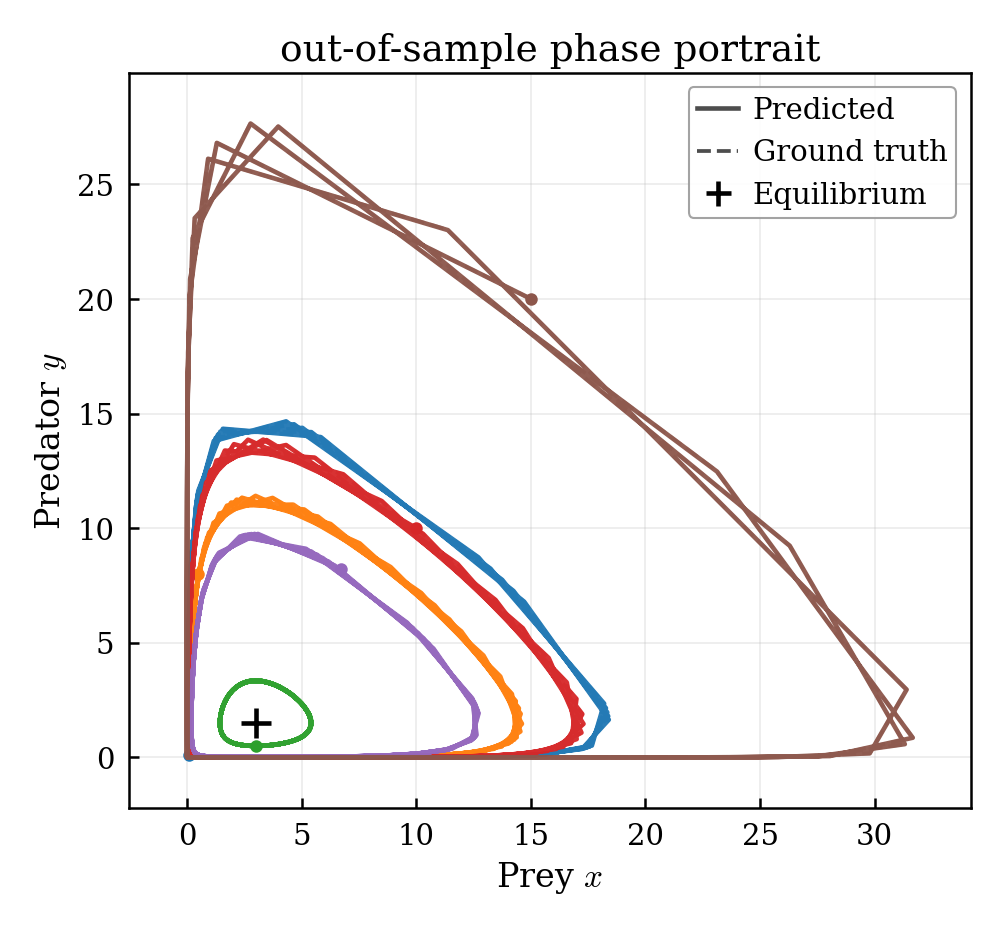} \\
\hline
\end{tabular}
\caption{\textbf{Per-method diagnostics: all five methods.} Each
row: in-sample time series, out-of-sample time series, out-of-sample
phase portrait. LV\_Structured keeps the tightest orbits; LV\_NN
spirals; \mpinode\ and LV\_MIC preserve orbital topology without
requiring full mechanistic knowledge.}
\label{fig:app_method_all}
\end{figure*}

%% ───────────────────────────────────────────────────────────
%%  APPENDIX E — Training Protocol Hyperparameters
%% ───────────────────────────────────────────────────────────
\section{Training Protocol Details}
\label{app:training-protocol}

This appendix collects the full training-protocol hyperparameter
recap referenced from \cref{sec:mpinode} and summarized in
\cref{alg:mpinode}. The same protocol is used for every method in
the comparison, so any observed difference reflects the algorithmic
ingredient under test rather than a protocol confound.

\begin{table}[tbp]
\centering
\small
\caption{\textbf{\mpinode\ training-protocol hyperparameters.}
Defaults used for every method in the comparison unless otherwise
noted. Method-specific weights (for example, $\lambda_\text{phys} = 0$
for LV\_NN/LV\_MIC and $\lambda_\text{cont} = 0$ for LV\_NN/LV\_PINN)
follow from the per-method ablation. Abbreviations: MSE
$=$ mean-squared error, MLP $=$ multilayer perceptron,
rtol $=$ relative tolerance, atol $=$ absolute tolerance.}
\label{tab:training_protocol}
\rowcolors{2}{gray!8}{white}
\begin{ruledtabular}
\begin{tabular}{ll}
Hyperparameter & Value \\
\colrule
Optimizer                              & Adam~\cite{Kingma2014Adam} \\
\addlinespace[1pt]
Learning-rate schedule                 & Cosine annealing \\
\addlinespace[1pt]
Initial learning rate                  & $3 \times 10^{-3}$ \\
\addlinespace[1pt]
Epochs                                 & $3000$ \\
\addlinespace[1pt]
Batch size (initial conditions / epoch) & $128$ \\
\addlinespace[1pt]
Multiple-shooting segments $K$         & $4$ \\
\addlinespace[1pt]
Physics weight $\lambda_\text{phys}$   & $10$ \\
\addlinespace[1pt]
Continuity weight $\lambda_\text{cont}$ & $1$ \\
\addlinespace[1pt]
$L_1$ weight-decay weight $\lambda_\text{reg}$ & $10^{-5}$ \\
\addlinespace[1pt]
Gradient-norm clip                     & $10$ \\
\addlinespace[1pt]
\addlinespace[1pt]
Integrator (evaluation)                & dopri5, rtol $=$ atol $= 10^{-8}$ \\
\addlinespace[1pt]
Architecture                           & $128 \times 4$ tanh MLP \\
\addlinespace[1pt]
Initialization                         & Xavier-normal weights, zero biases \\
\addlinespace[1pt]
Positivity wrapper                     & Clamp $[-20, 20]$ \\
\addlinespace[1pt]
IC distribution $D_\text{IC}$          & 50/50 mix: uniform $[0.1, 10.0]^2$ \\
                                       & and log-uniform $[10^{-3}, 10.0]^2$ \\
\addlinespace[1pt]
Validation set                         & $64$ held-out ICs from same mix \\
\addlinespace[1pt]
Checkpointing                          & best validation MSE retained \\
\end{tabular}
\end{ruledtabular}
\end{table}

\clearpage
\bibliography{References}

\end{document}